\documentclass[11pt]{article}

\usepackage[preprint]{acl}

\usepackage{times}
\usepackage{latexsym}
\usepackage{booktabs}
\usepackage{multirow}
\usepackage{colortbl}
\usepackage{xcolor}
\usepackage{cleveref}
\usepackage{amssymb}
\usepackage{parskip}
\usepackage{tcolorbox}
\usepackage{todonotes}

\usepackage{graphicx}
\usepackage{subcaption}
\usepackage{longtable}
\usepackage{array}    
\usepackage{booktabs} 
\usepackage[T1]{fontenc}
\usepackage{numprint}
\npthousandsep{,}

\usepackage[utf8]{inputenc}

\newcommand\datasetname{\textcolor{black}{\textsc{POLAR}}}

\newcommand\polardetect{\textcolor{black}{\textbf{\textsc{PolarDetect}}}}
\newcommand\polartype{\textcolor{black}{\textbf{\textsc{PolarType}}}}
\newcommand\polarmanifest{\textcolor{black}{\textbf{\textsc{PolarManifest}}}}

\usepackage{microtype}
\usepackage{inconsolata}
\usepackage{graphicx}
\usepackage{enumitem}
\definecolor{lightblue}{RGB}{204, 229, 255}
\definecolor{lightorange}{RGB}{255, 229, 204}

\title{\datasetname: A Benchmark for Multilingual, Multicultural, \\ and Multi-Event Online Polarization}

\author{
	Usman Naseem$^{1}$,
	Robert Geislinger$^{2}$,
	Juan Ren$^{1}$,
	Sarah Kohail$^{3}$,
	\\
	\bf Rudy Garrido Veliz$^{2}$, P Sam Sahil$^{2,4}$, Yiran Zhang$^{1}$, Marco Antonio Stranisci$^{6,7}$,
    \\
    \bf Idris Abdulmumin$^{5}$, Özge Alacam$^{8}$, Cengiz Acartürk$^{9}$, Aisha Jabr$^{3}$,Saba Anwar$^{2}$, 
    \\
	\bf Abinew Ali Ayele$^{10}$, Simona Frenda$^{7,11}$, Alessandra Teresa Cignarella$^{7,}$
    \\
	\bf Elena Tutubalina$^{13,14,15}$, Oleg Rogov$^{13,16,17}$, Aung Kyaw Htet$^{1}$, Xintong Wang$^{2}$,
    \\
	\bf Surendrabikram Thapa$^{18}$, Kritesh Rauniyar$^{1}$, Tanmoy Chakraborty$^{19}$,
    \\
	\bf Arfeen Zeeshan$^{19}$, Dheeraj Kodati$^{20}$, Satya Keerthi$^{21}$, Sahar Moradizeyveh$^{1}$,
    \\
	\bf Firoj Alam$^{22,23}$, Arid Hasan$^{24}$, Syed Ishtiaque Ahmed$^{24}$, Ye Kyaw Thu$^{25}$,
	\\
	\bf Shantipriya Parida$^{26}$,
	Ihsan Ayyub Qazi$^{27}$,
	Lilian Wanzare$^{28}$,
    \\
	\bf Nelson Odhiambo Onyango$^{28}$,
	Clemencia Siro$^{29}$,
	Jane Wanjiru Kimani$^{30}$,
    \\
	\bf Ibrahim Said Ahmad$^{31,32}$,
	Adem Chanie Ali$^{2,10}$,
	Martin Semmann$^{2}$,
    \\
	\bf Chris Biemann$^{2}$,
	Shamsuddeen Hassan Muhammad$^{33}$,
	Seid Muhie Yimam$^{2}$
    \\
	\footnotesize $^{1}$Macquarie University,
	\footnotesize $^{2}$University of Hamburg,
	\footnotesize $^{3}$Zayed University,
    \footnotesize $^{4}$HKBK College of Engineering,
	\\
	\footnotesize $^{5}$University of Pretoria,
    \footnotesize $^{6}$University of Turin,
    \footnotesize $^{7}$aequa-tech,
    \footnotesize $^{8}$Bielefeld University,
	\footnotesize $^{9}$Jagiellonian University,
    \\
	\footnotesize $^{10}$Bahir Dar University,
	\footnotesize $^{11}$Heriot-Watt University,
	\footnotesize $^{12}$Ghent University,
    \footnotesize $^{13}$AIRI,
    \footnotesize $^{14}$KFU,
    \footnotesize $^{15}$HSE University,
    \\
    \footnotesize $^{16}$MTUCI,
    \footnotesize $^{17}$Skoltech,
    \footnotesize $^{18}$Virginia Tech,
    \footnotesize $^{19}$ITT Delhi,
    \footnotesize $^{20}$ABV-IIITM,
    \footnotesize $^{21}$Mahindra University,
    \\
    \footnotesize $^{22}$Qatar Computing Research Institute, 
    \footnotesize $^{23}$Hamad Bin Khalifa University,
    \footnotesize $^{24}$University of Toronto,
    \\
    \footnotesize $^{25}$LU Lab., Myanmar,
    \footnotesize $^{26}$AMD Silo AI,
    \footnotesize $^{27}$Lahore University of Management Sciences,
    \footnotesize $^{28}$Maseno University,
    \\
    \footnotesize $^{29}$Centrum Wiskunde \& Informatica,
    \footnotesize $^{30}$Jomo Kenyatta University of Agriculture and Technology,
    \\
	\footnotesize $^{31}$Bayero University Kano,
	\footnotesize $^{32}$Northeastern University,
	\footnotesize $^{33}$Imperial College London,
    \\
}

\begin{document}
\maketitle
\begin{abstract}

Online polarization poses a growing challenge for democratic discourse, yet most computational social science research remains monolingual, culturally narrow, or event-specific. 
We introduce \datasetname{}, a multilingual, multicultural, and multi-event dataset with over 110K instances in 22 languages drawn from diverse online platforms and real-world events. 
Polarization is annotated along three axes, namely \textit{detection}, \textit{type}, and \textit{manifestation}, using a variety of annotation platforms adapted to each cultural context. 
We conduct two main experiments: \textit{(1)} fine-tuning six pretrained small language models; and \textit{(2)} evaluating a range of open and closed large language models in few-shot and zero-shot settings. 
The results show that, while most models perform well in binary polarization detection, they achieve substantially lower performance when predicting polarization types and manifestations. 
These findings highlight the complex, highly contextual nature of polarization and demonstrate the need for robust, adaptable approaches in NLP and computational social science. 
All resources will be released to support further research and effective mitigation of digital polarization globally.
\end{abstract}

\section{Introduction}

Online polarization, defined as sharp division and antagonism between social, political, or identity groups, has become a pervasive threat to democratic institutions, civil discourse, and social cohesion worldwide~\citep{waller2021quantifying}.
It is often fueled by biased or inflammatory content in digital media, strengthening echo chambers and undermining mutual understanding~\citep{garimella2018polarization}.
Polarized discourse amplifies ideological divides and can escalate into hate speech, harassment, and real-world violence \cite{piazza2023political,martinez2024methodology}.
Therefore, early detection of polarization is essential for designing interventions that promote healthier online ecosystems.

\begin{figure}[!t]
    \centering
    \includegraphics[width=1.0\linewidth]{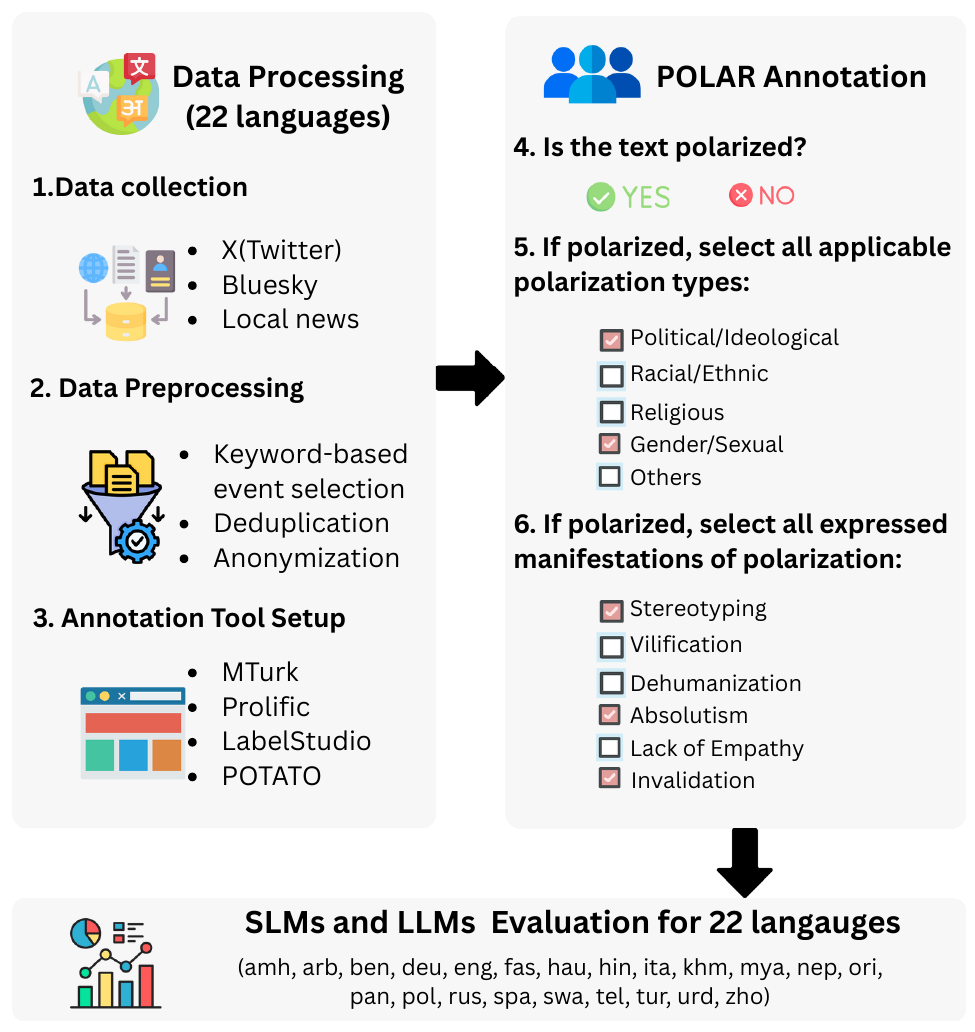}
    \vspace{-0.35cm}
    \caption{\textbf{Pipeline for \datasetname~construction.} Data curation in 22 languages, annotation workflow, and benchmarking.}
    \label{fig:pipeline}
    \vspace{-0.35cm}
\end{figure}

Despite increasing attention, computational approaches to polarization face several limitations.
\textit{First}, most existing datasets focus on English or high-resource languages, reflecting a widespread trend across NLP tasks that ignores the rich diversity of linguistic and sociocultural contexts in which polarization manifests \cite{simchon2022troll, piazza2023political, rojo2025affective}.
\textit{Second}, previous studies are event-specific or monodomain, such as U.S. elections or Western political debates, limiting their generalizability \cite{demszky-etal-2019-analyzing,casal2021polarization,sinno-etal-2022-political, piazza2023political}.
\textit{Third}, the conceptualization of polarization in NLP has largely been binary or topic-focused \cite{hofmann-etal-2022-modeling}, overlooking the multifaceted ways in which polarization is expressed through vilification, dehumanization, stereotyping, or other rhetorical tactics \cite{donohue2022framework}.
These tactics are often employed in political rhetoric, social debates, or campaigns to solidify support within a group and increase hostility to others.

\begin{figure}
    \centering
    \includegraphics[width=1\linewidth]{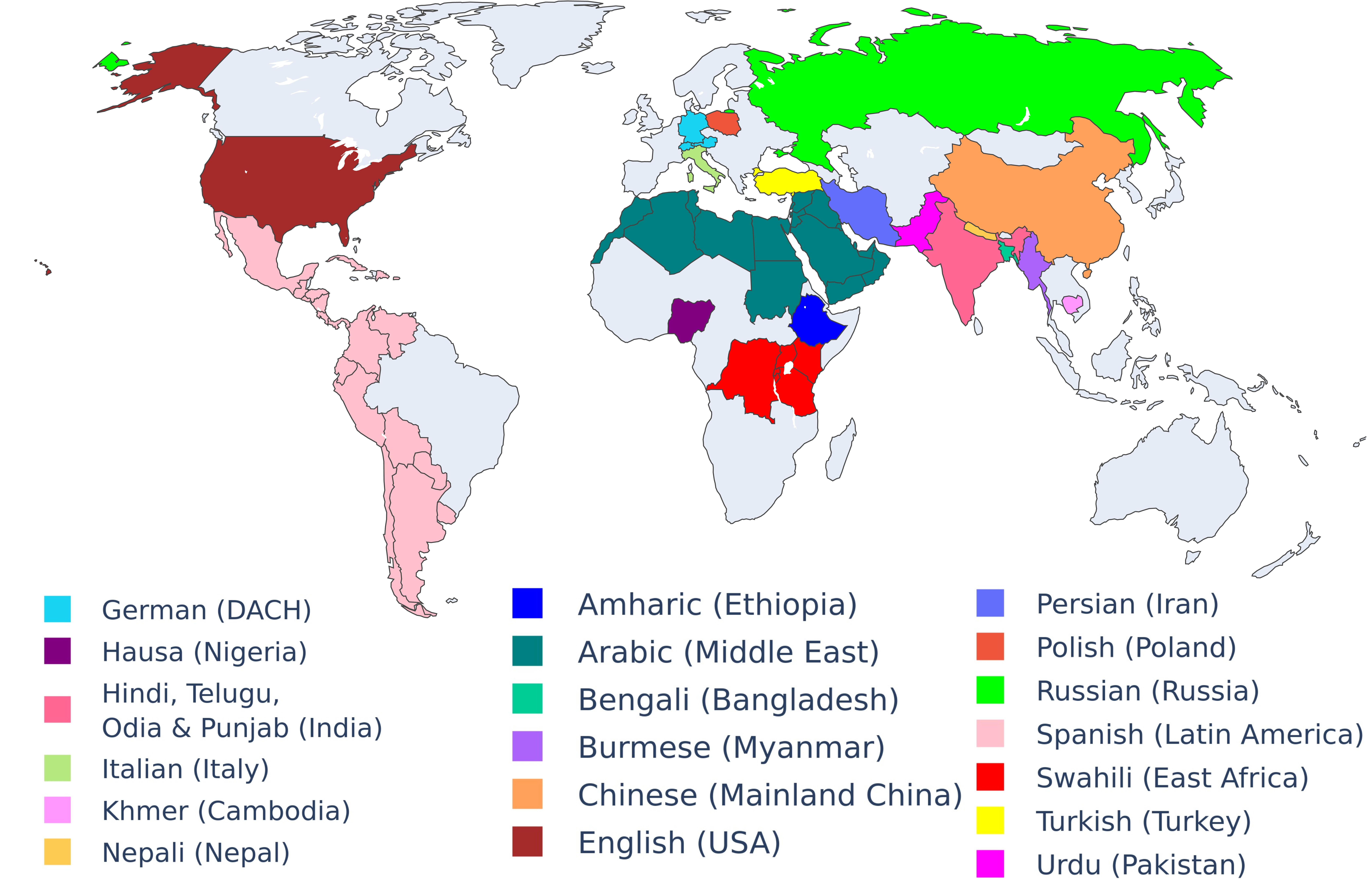}
    \caption{Languages represented in a world map covered by \datasetname, covering diverse linguistic and regional contexts. The language and societal context can present itself across varied areas. Language assignments to countries and regions are approximate.}
    \label{fig:lang-map}
    \vspace{-5mm}

\end{figure}

To address these gaps, we introduce \datasetname, a large-scale, multilingual, multicultural, and multievent dataset for fine-grained polarization detection. \datasetname~supports 22 languages spanning seven language families and balances high-, medium-, and low-resource languages (see \Cref{tab:language_family}). The wide extent of our efforts can be seen in Figure \ref{fig:lang-map}. Unlike prior work, \datasetname~supports three complementary tasks:


\begin{itemize}[noitemsep,leftmargin=*]
   
\item \textbf{Binary Polarization Detection:}  Determine whether a given text expresses polarization. We refer to this task as  \polardetect. 

 \item \textbf{Polarization Type Classification:} Identify the social dimension underlying the polarization
 (e.g., political, religious, racial). We refer to this task as \polartype. 

\item \textbf{Manifestation Identification:} Detect how polarization is rhetorically manifested, including strategies such as stereotyping, deindividuation, vilification, dehumanization, extreme language, and other rhetorical devices. We refer to this task as \polarmanifest.

\end{itemize}
 
For each task, we develop a cross-cultural annotation protocol tailored for each language's sociopolitical context. The complete data construction pipeline is illustrated in \Cref{fig:pipeline}. We benchmark a range of Small Language Models (SLMs) and Large Language Models (LLMs) under zero-shot and few-shot settings. Our experiments highlight the challenges of generalization and the limitations of current models in capturing nuanced rhetorical patterns across languages. Our contributions are as follows: 

\begin{itemize}[noitemsep,leftmargin=*]
\item We release \datasetname, the first large-scale, multilingual, fine-grained dataset for polarization analysis across 22 languages and diverse global events, comprising 110K instances.
  \item We define a taxonomy of polarization types and manifestations, operationalized through a robust cross-lingual annotation protocol.
\item We provide comprehensive benchmarks using state-of-the-art SLMs and LLMs across multiple evaluation settings.
\end{itemize}

\section{Related Work}

Online polarization poses a threat to social cohesion, exacerbated by social media echo chambers and biased content~\citep{waller2021quantifying, iandoli2021impact, garimella2018polarization}. As social media and other online platforms become key arenas for political and cultural discourse, the need for early detection and nuanced understanding of polarization has grown significantly. Polarization detection is important for content moderation, peace building, policy development, responsible digital governance and healthy democracy. Foundational research has defined polarization as both intergroup hostility and blind ingroup cohesion~\citep{arora2022polarization}, and has highlighted its relationship with hate speech, fragmentation, and incivility~\citep{ mathew2020hatexplain}.

A growing body of research has documented the role of online spaces in intensifying polarization across different regions \citep{kubin2021role, barbera2020social, gitlin2016outrage, soares2021hashtag}. However, most computational work focuses on high-resource languages and event- or region-specific datasets, limiting generalizability~\citep{kubin2021role}. This leaves a significant gap in our ability to generalize findings across cultures, languages, and events, especially for low-resource languages or multilingual regions.

The lack of standardized datasets across languages has hindered progress in developing and evaluating polarization detection models with cross-lingual or cross-cultural capabilities. Recent shared tasks on hate speech and toxicity~\citep{basile2019semeval, pamungkas2020misogynistic} have expanded the language and domain coverage, yet remain less fine-grained regarding polarization’s diverse types and rhetorical manifestations. Our work addresses this gap by presenting the comprehensive, fine-grained dataset benchmark for multilingual, multicultural, and multievent online polarization, enabling robust cross-lingual and context-aware modeling.

\section{\datasetname~Dataset Construction}



\subsection{Operational Definitions}
\label{polardef}

In this work, we define polarization as the increasing extremity of opinions, beliefs, or behaviors, resulting in heightened intergroup divisions and conflict. Polarization types are classified as:

\begin{itemize}[noitemsep,leftmargin=*]
    \item \textbf{Political polarization}: Focuses on division, intolerance, and conflict between political parties and followers. 
    \item \textbf{Racial or ethnic polarization}: Focuses on ethnic identity or racial origin and incites division, intolerance, and conflict between ethnic groups or races. 
    \item \textbf{Religious  polarization:} Focuses on religious identity and incites division, intolerance, and conflict between religious followers
    \item \textbf{Gender/ Sexual polarization}: Refers to the exclusion, discrimination, and marginalization of individuals based on their gender or sexual orientations.

    \item \textbf{Other}: Polarized texts targeting other groups or identities not covered above, such as economic class, technology or media.

\end{itemize}

In addition to topical categories, we further distinguish polarization by its rhetorical manifestations, defined as follows:

\begin{itemize}[noitemsep,leftmargin=*]
    \item \textbf{Stereotype}: A generalized belief that attributes specific characteristics to all members of a group, often neglecting individual differences, thereby reducing complex personalities to simplistic and uniform representations.
    \item  \textbf{Vilification}: The act of defaming or demonizing a particular group, person, or entity by inciting fear, often through exaggeration, misrepresentation, or biased framing that portrays the subject negatively and harmfully.
    \item \textbf{Dehumanization}: The process of depriving a group or individual of their human qualities or personality by comparing them to animals, machines, or objects, or otherwise denying their humanity, dignity, or individuality.
    \item \textbf{Extreme Language and Absolutism}: The use of language that is extreme or makes definitive, all-encompassing statements, often involving words like ``always'', ``never'', ``worst'', or ``best'', and presenting issues in a dichotomous manner such as ``us versus them'' or ``right versus wrong''.
    \item  \textbf{Lack of Empathy}: The absence of compassion or recognition for other viewpoints or experiences in the text.
    \item \textbf{Invalidation}: The act of denying or dismissing the identity and existence of individuals or groups, thereby rejecting their sense of self and their presence.
\end{itemize}
Appendix~\ref{sec:Annotation_Guidelines} contains more details and examples for each manifestation of polarization.\footnote{Since we define the above polarization manifestations as rhetorical tactics, we have used the terms ``manifestation'' and ``rhetorical tactics'' interchangeably.}

\subsection{Data Collection}



\textbf{Data Sources:} We collected data from a range of online platforms, including major social media sites (e.g., X, Facebook, Reddit, Bluesky, Threads, YouTube comments, Weibo, and Zhihu) and local news or commentary forums (see Table~\ref{tab:data_sources} in the Appendix).
For several languages (Chinese, Turkish, Polish, Burmese, and Italian), we sampled and re-annotated instances from existing toxic or hate speech datasets, including ToxiCN~\citep{bai2025statetoxicnbenchmarkspanlevel}, COLD~\citep{deng-etal-2022-cold}, the Turkish Hate Speech Dataset~\citep{coltekin-2020-corpus}, Myanmar Hate Speech~\citep{10613636}, HaSpeede2~\citep{Sanguinetti:2020}, HODI~\citep{Nozza:2023}, and BAN-PL~\citep{kołos2024banpl}.




\textbf{Event Selection:} We curated the dataset to cover diverse real-world events, grounding event selection in the sociopolitical and socioeconomic contexts specific to each language and cultural setting. The data span a broad range of events and issues, including armed conflicts (e.g., the Tigray War in Ethiopia, the Russia–Ukraine conflict, and the Gaza genocide), elections and party politics (e.g., the 2024 U.S. and 2025 German elections), public health crises, large-scale migration, climate change, and broader socioeconomic debates. The dataset also includes discussions related to gender and indigenous rights, religion, and ideology. For some languages, such as Bengali, a broader sampling strategy was adopted due to the lack of sufficiently large event-specific data on the selected platforms.

\begin{table*}[!t]
\centering
\scriptsize
\setlength{\tabcolsep}{3.2pt}
\renewcommand{\arraystretch}{1}
\resizebox{\textwidth}{!}{%
\begin{tabular}{l|l| c | r |r| rrrrr| rrrrrr}
\toprule
\multirow{3}{*}{\#}&
\multirow{3}{*}{\textbf{Lang.}} &
\multirow{3}{*}{\textbf{\shortstack{Inner \\ Agr. ($\kappa$)}}} &
\multirow{3}{*}{\textbf{Total}} &
\multicolumn{1}{c}{\polardetect} &
\multicolumn{5}{c}{\polartype} &
\multicolumn{6}{c}{\polarmanifest} \\
\cmidrule(lr){5-5}
\cmidrule(lr){6-10}
\cmidrule(lr){11-16}

& & & &
\textbf{Polarized (\%)} &
\textbf{Political} &
\textbf{\shortstack{Racial / \\ Ethnic}} &
\textbf{\shortstack{Religious \\ Polarization}} &
\textbf{\shortstack{Gender / \\ Sexual}} &
\textbf{Other} &
\textbf{\shortstack{Stereo- \\ type}} &
\textbf{\shortstack{Vilifi- \\ cation}} &
\textbf{\shortstack{Dehuman- \\ ization}} &
\textbf{\shortstack{Extreme \\ Language}} &
\textbf{\shortstack{Lack of \\ Empathy}} &
\textbf{\shortstack{Invalid- \\ ation}} \\
\midrule

\multirow{13}{*}{1}
& eng & 0.39 & 4,834 & 1,767 (37\%) & 1,726 & 422 & 168 & 108 & 190 & 730 & 1,272 & 586 & 1,156 & 536 & 879 \\
& deu & 0.10$^{*}$ & 4,771 & 2,274 (48\%) & 1,959 & 883 & 531 & 281 & 658 & 1,728 & 1,435 & 712 & 1,038 & 1,272 & 775 \\
\cmidrule{2-16}
& urd & 0.29 / 0.70$^{*}$ & 5,346 & 3,714 (69\%)& 3,603 & 2,908 & 2,954 & 2,739 & 2,713 & 3,328 & 3,460 & 2,973 & 3,324 & 3,007 & 3,059 \\
& ben & 0.59 & 5,000 & 2,127 (43\%)& 1,701 & 38 & 97 & 26 & 503 & 298 & 1,199 & 535 & 236 & 95 & 89 \\
& hin & 0.49 & 4,117 & 3,510 (85\%)& 3,051 & 500 & 2,417 & 472 & 540 & 2,047 & 2,683 & 750 & 2,082 & 2,336 & 2,703 \\
& ori & 0.46 & 3,552 & 1,021 (29\%)& 744 & 179 & 225 & 119 & 130 & 354 & 385 & 24 & 476 & 56 & 120 \\
& nep & 0.79 & 3,008 & 1,510 (50\%)& 518 & 422 & 239 & 158 & 354 & 806 & 947 & 198 & 816 & 318 & 450 \\
& pan & 0.55$^{*}$ & 2,609 & 1,280 (49\%)& 803 & 153 & 205 & 291 & 233 & 424 & 1,038 & 574 & 624 & 324 & 637 \\
\cmidrule{2-16}
& ita & 0.39 & 5,038 & 2,165 (44\%)& 412 & 926 & 368 & 461 & 219 & \multicolumn{6}{c}{--} \\
& spa & 0.26 & 4,958 & 2,479 (50\%)& 1,351 & 945 & 787 & 665 & 665 & 1,355 & 1,517 & 443 & 1,199 & 1,187 & 526 \\
\cmidrule{2-16}
& rus & 0.39 & 5,023 & 1,525 (30\%)& 696 & 494 & 205 & 284 & 119 & \multicolumn{6}{c}{--} \\
& pol & 0.46 & 3,587 & 1,504 (42\%)& 1,313 & 323 & 131 & 165 & 232 & \multicolumn{6}{c}{--} \\
\cmidrule{2-16}
& fas & 0.78 & 4,943 & 3,656 (74\%)& 2,170 & 120 & 476 & 296 & 1,197 & 649 & 2,850 & 213 & 835 & 487 & 394 \\

\midrule
\multirow{3}{*}{2}
& hau & 0.48 & 5,477 & 587 (11\%)& 267 & 173 & 139 & 44 & 21 & 234 & 68 & 193 & 165 & 48 & 13 \\
& arb & 0.25 & 5,070 & 2,268 (45\%)& 1,205 & 874 & 424 & 553 & 847 & 1,691 & 1,896 & 555 & 1,540 & 863 & 411 \\
& amh & 0.59 & 4,999 & 3,747 (75\%)& 3,339 & 1,296 & 99 & 29 & 1,239 & 2,728 & 2,398 & 657 & 1,527 & 879 & 799 \\

\midrule
\multirow{2}{*}{3}
& zho & 0.64 & 6,421 & 3,208 (50\%)& 376 & 1,475 & 127 & 1,085 & 552 & 1,931 & 1,188 & 323 & 522 & 506 & 307 \\
& mya & 0.13 & 4,334 & 2,508 (58\%)& 1,095 & 228 & 133 & 459 & 1,956 & \multicolumn{6}{c}{--} \\

\midrule
4 & swa & 0.56 & 10,487 & 5,257 (50\%)& 279 & 3,721 & 371 & 234 & 833 & 4,160 & 4,324 & 1,340 & 2,509 & 3,120 & 2,456 \\

\midrule
5 & khm & 0.83 & 9,960 & 9,042 (91\%)& 1,825 & 147 & 336 & 169 & 6,565 & 6,799 & 152 & 122 & 225 & 1,093 & 651 \\

\midrule
6 & tel & 0.70 & 3,550 & 1,885 (53\%)& 766 & 603 & 318 & 471 & 842 & 398 & 781 & 88 & 477 & 933 & 809 \\

\midrule
7 & tur & 0.46 & 3,566 & 1,776 (50\%)& 1,569 & 579 & 557 & 221 & 193 & 1,453 & 1,169 & 378 & 1,575 & 369 & 159 \\

\midrule
\multicolumn{3}{c}{\textbf{Total}} &
\textbf{110,650} & \textbf{58,810 (53\%)} &
\textbf{30,768} & \textbf{17,409} & \textbf{11,307} &
\textbf{9,330} & \textbf{20,801} &
\textbf{31,113} & \textbf{28,762} &
\textbf{10,664} & \textbf{20,326} &
\textbf{17,429} & \textbf{15,237} \\
\bottomrule
\end{tabular}%
}
\vspace{-0.25cm}
\caption{Number of samples labeled as positive for each annotation task across languages. Inner agreement values denote inter-annotator agreement per language (Fleiss’s $\kappa$ unless otherwise noted). $^{*}$ denotes exceptions: German uses Krippendorff’s $\alpha$; Punjabi reports identical Krippendorff’s $\alpha$ and Cohen’s $\kappa$; Urdu reports Fleiss’s $\kappa$ / Cohen’s $\kappa$. Polarization manifestation annotations are not available for Italian, Russian, Burmese, and Polish. Languages are ordered by language families and sub-branches as defined in Table~\ref{tab:language_family}.}
\label{tab:label_prevalence}
\vspace{-0.55cm}
\end{table*}


To collect data for these events, we adopted a dynamic, keyword-driven strategy tailored to each language and topic. Keyword lists were curated by human experts and native speakers to capture culturally and politically salient discourse and were used to retrieve data from online platforms. Table~\ref{tab:dataset_composition} in the Appendix provides an overview of the dataset composition, size, and event coverage.



\textbf{Dataset Quality Control:} To ensure high-quality annotations across languages, we implemented several steps to ensure data quality throughout the quality control process.
Before annotation, native speakers developed language-specific preprocessing pipelines.
These pipelines included standard NLP procedures such as tokenization, word count filtering, and duplicate detection.
Instances that were either too short or excessively long based on language-specific thresholds were removed.
For anonymization, all usernames and URLs were replaced with standardized placeholders.
For some languages, LLMs were used during the pre-filtering stage to increase the proportion of polarized content. 

\subsection{Annotation Process and Guidelines}


We used a hybrid annotation strategy, leveraging crowd-sourced annotators and trained community annotators for low-resource languages where crowd-sourced annotation support is limited. For the crowd-sourced setting, annotators were selected based on their prior experience and annotation quality. Specifically, we filtered candidates using historical annotation agreement scores and conducted pilot rounds to identify those with consistent performance. Only annotators achieving a Fleiss’ Kappa score of at least 0.8 were retained for the main task. For community annotation, we recruited native speakers with at least a bachelor’s degree. Annotators received training, followed by a pilot round to assess their understanding and performance. Subsequent annotation was assigned in batches, with annotator performance monitored continuously to ensure consistency and adherence to guidelines.

The trained annotators used POTATO~\cite{pei-etal-2022-potato} and Label Studio\footnote{\url{https://labelstud.io}} as annotation platforms, while the crowd-sourced annotators used Mechanical Turk\footnote{\url{https://www.mturk.com}} and Prolific\footnote{\url{https://www.prolific.com}}.

\noindent\textbf{Annotation Guidelines:} Given the cultural and linguistic breadth of \datasetname, we developed detailed, multilingual annotation guidelines (see Appendix~\ref{sec:Annotation_Guidelines}) in English, and then translated and culturally adapted them for each target language. 
Annotators were instructed to:
\begin{itemize}[noitemsep,leftmargin=*]
 \item Identify whether a text is polarized
 \item If the text is classified as polarized, tag the type of polarization (political, racial/ethnic, religious, gender/sexual identity, other)
 \item If the text is classified as polarized, tag its manifestations or rhetorical tactics (stereotyping/deindividuation, vilification, dehumanization, extreme language, lack of empathy, invalidation).
\end{itemize}

Multiple labels were allowed due to the conceptual and contextual overlap often observed in polarized content.

\subsection{Annotators’ Reliability} 

To evaluate annotation quality, we report Fleiss’ Kappa as the inter-annotator agreement (IAA) metric. As shown in \Cref{tab:label_prevalence}, the IAA scores vary between languages, with the majority showing moderate agreement and a few, such as "khm" and "tel" achieving good agreement. 
Although guidelines were standardized, their interpretation was influenced by cultural and political context, especially in languages with lower agreement, where some terms may not have direct equivalents across cultures. 
 Latent content or sarcasm often required annotators to draw on their own socio-political knowledge, highlighting the perspectivist nature of polarization \citep{cabitza2023toward}. Thus, low agreement can indicate socio-pragmatic complexity rather than error, signaling that polarization markers may not have universal meanings and that divergences can reveal inherent ambiguity in stimuli or interpretation \citep{aroyo2015truth}. Examples illustrating such ambiguities are provided below.

\begin{tcolorbox}
\textbf{Unanimous polarized:} \textit{``However I got tanned in Bodrum and I don’t wait in line at the hospital.''} \\ This sentence was rated as polarized by all three annotators with high confidence. Here, the sentence refers to perceived privileges granted to the immigrant community.
\end{tcolorbox}

\begin{tcolorbox}
\textbf{Non-unanimous polarized}: \textit{``Do you agree with the idea that new citizens in our country shouldn’t be allowed to vote for ten years?''} This sentence did not receive consensus among annotators, leaning toward being polarized. The topic mentioned here is more complex and sensitive compared to the previous example.
\end{tcolorbox}

\subsection{Dataset Statistics}

A comprehensive analysis of the dataset was conducted using the annotated labels to support systematic examination. Table~\ref{tab:label_prevalence} provides a quantitative breakdown of positive labels for each annotation task. For further details on data sources, language-wise composition, and polarization statistics, including distributions of polarized instances, types, and manifestations, see Appendix~\ref{app:data_stats} (Tables~\ref{tab:data_sources}–\ref{tab:dataset_composition} and Figure~\ref{fig:polarity_manifestion}).
Sample data is shown in Table~\ref{tab:sample_data}, \ref{tab:SampleGerman}, and \ref{tab:SampleSpanish}.

\begin{table*}[!t]
\centering
\small
\setlength{\tabcolsep}{3pt}
\renewcommand{\arraystretch}{1}
\resizebox{\textwidth}{!}{%
\begin{tabular}{l|l|cccccc|cccccc|ccccccc}
\toprule
\multirow{2}{*}{\#} &
\multirow{2}{*}{\textbf{Lang.}} &
\multicolumn{6}{c}{\polardetect} &
\multicolumn{6}{c}{\polartype} &
\multicolumn{6}{c}{\polarmanifest} \\
\cmidrule(lr){3-8}\cmidrule(lr){9-14}\cmidrule(lr){15-20}
& & \textbf{mBERT} & \textbf{XLM-R} & \textbf{RemBERT} & \textbf{LaBSE}&
\shortstack{\textbf{twitter-}\\\textbf{roberta-hate}} & \shortstack{\textbf{afro-xlmr-}\\\textbf{large}}
& \textbf{mBERT} & \textbf{XLM-R} & \textbf{RemBERT} & \textbf{LaBSE} & 
\shortstack{\textbf{twitter-}\\\textbf{roberta-hate}} & \shortstack{\textbf{afro-xlmr-}\\\textbf{large}}
& \textbf{mBERT} & \textbf{XLM-R} & \textbf{RemBERT} & \textbf{LaBSE} & 
\shortstack{\textbf{twitter-}\\\textbf{roberta-hate}} & \shortstack{\textbf{afro-xlmr-}\\\textbf{large}}
\\
\midrule
\multirow{13}{*}{1}
& eng & 74.83 & 76.34 & \cellcolor{lightorange}77.51 & 77.27 & \cellcolor{lightblue}79.59 & 75.54 &
      31.29 & 27.33 & \cellcolor{lightblue}43.17 & 34.68 & \cellcolor{lightorange}36.99 & 19.09 & 
      41.83 & 44.50 & \cellcolor{lightorange}47.90 & 46.58 & \cellcolor{lightblue}49.04 & 47.82 \\
& deu & 65.69 & \cellcolor{lightorange}69.03 & \cellcolor{lightblue}69.81 & 68.57 & 65.97 & 59.14 & 
      49.34 & 48.89 & \cellcolor{lightblue}54.23 & 53.26 & 49.39 & \cellcolor{lightorange}51.34 & 
      43.74 & 45.24 & \cellcolor{lightblue}47.46 & \cellcolor{lightorange}47.13 & 44.28 & 45.18 \\
\cmidrule{2-20}
&urd & 71.60 & 68.74 & \cellcolor{lightblue}75.62 & \cellcolor{lightorange}74.20 & 63.45 & 73.51 
    & 68.76 & 72.16 & \cellcolor{lightblue}72.96 & \cellcolor{lightorange}73.91 & 71.05 & 67.16 
    & 72.13 & \cellcolor{lightorange}72.83 & 74.75 & \cellcolor{lightblue}77.09 & 72.49 & 43.00 \\
&ben & 79.03 & \cellcolor{lightblue}83.68 & 82.74 & 82.50 & 36.81 & \cellcolor{lightorange}83.04 & 
      \cellcolor{lightblue}24.61 & 16.59 & 10.14 & \cellcolor{lightblue}26.82 & 10.14 & 13.62 & 
      21.08 & \cellcolor{lightblue}24.12 & 5.05 & \cellcolor{lightblue}25.83 & 9.59 & 5.64 \\
&hin & \cellcolor{lightorange}76.37 & 75.06 & 45.98 & \cellcolor{lightblue}78.23 & 61.80 & 65.49 & 
      61.40 & 63.58 & \cellcolor{lightblue}73.30 & \cellcolor{lightorange}69.97 & 38.83 & 57.87 & 
      53.71 & 69.12 & \cellcolor{lightblue}72.69 & \cellcolor{lightorange}70.13 & 63.58 & 62.07 \\
&ori & 41.72 & 41.72 & \cellcolor{lightorange}71.14 & \cellcolor{lightblue}77.60 & 42.42 & 68.73 
    & 19.44 & 28.14 & \cellcolor{lightorange}32.92 & \cellcolor{lightblue}42.27 & 19.23 & 29.47 
    & 8.37 & 4.84 & 2.78 & \cellcolor{lightblue}24.00 & 5.77 & \cellcolor{lightorange}9.47 \\
&nep & \cellcolor{lightorange}85.04 & 84.04 & 87.48 & \cellcolor{lightblue}88.26 & 69.86 & 76.09 
    & 45.75 & 48.78 & \cellcolor{lightblue}70.53 & \cellcolor{lightorange}66.42 & 32.69 & 49.16 
    & 55.15 & 53.92 & \cellcolor{lightorange}58.11 & \cellcolor{lightblue}60.19 & 39.19 & 35.55 \\
&pan & 69.98 & 66.50 & \cellcolor{lightorange}73.57 & \cellcolor{lightblue}74.91 & 61.03 & 62.28  
    & 29.59 & 28.22 & \cellcolor{lightblue}40.88 & \cellcolor{lightorange}38.73 & 27.32 & 29.08 
    & 42.07 & 40.07 & \cellcolor{lightorange}45.80 & \cellcolor{lightblue}48.35 & 38.57 & 43.31 \\
\cmidrule{2-20}
&spa & 70.36 & 71.27 & \cellcolor{lightblue}75.24 & 74.99 & 71.96 & \cellcolor{lightorange}74.24 
    & 58.22 & 54.54 & \cellcolor{lightblue}65.01 & \cellcolor{lightorange}59.34 & 57.78 & 58.64 
    & 41.17 & 43.34 & \cellcolor{lightblue}50.63 & \cellcolor{lightorange}48.04 & 43.76 & 46.45 \\
&ita & 54.35 & 34.50 & \cellcolor{lightblue}60.93 & \cellcolor{lightorange}56.40 & 53.00 & 47.78 & 
      21.48 & 21.83 & \cellcolor{lightorange}25.99 & \cellcolor{lightblue}26.13 & 21.08 & 23.47 & 
      -- & -- & -- & -- & -- & -- \\
\cmidrule{2-20}
&rus & 70.20 & 71.61 & \cellcolor{lightblue}77.39 & 74.77 & 42.33 & \cellcolor{lightorange}74.80  
    & 31.97 & 21.93 & \cellcolor{lightblue}47.33 & \cellcolor{lightorange}40.94 & 15.81 & 36.44 
    & -- & -- & -- & -- & -- & -- \\
&pol & 74.45 & 74.54 & \cellcolor{lightblue}77.96 & \cellcolor{lightorange}77.26 & 68.45 & 67.62 
    & 31.63 & 31.29 & \cellcolor{lightblue}46.84 & \cellcolor{lightorange}41.55 & 24.65 & 24.48 
    & -- & -- & -- & -- & -- & -- \\
\cmidrule{2-20}
&fas & 80.09 & 78.34 & 77.50 & \cellcolor{lightblue}83.50 & 54.64 & \cellcolor{lightorange}80.89 &
      46.69 & 47.21 & \cellcolor{lightblue}54.18 & \cellcolor{lightorange}52.46 & 27.72 & 27.75 & 
      35.53 & 34.42 & \cellcolor{lightblue}41.35 & \cellcolor{lightorange}39.51 & 24.72 & 34.78 \\
\midrule
\multirow{3}{*}{2} & hau & \cellcolor{lightblue}82.97 & 47.19 & 75.91 & \cellcolor{lightorange}82.09 & 78.32 & 81.82 & 
    \cellcolor{lightorange}21.44 & 15.75 & 15.53 & \cellcolor{lightblue}21.58 & 16.68 & 2.76 & 
     19.97 & 16.43 & \cellcolor{lightorange}21.44 & 20.55 & 18.29 & \cellcolor{lightblue}22.16 \\
&arb & 78.24 & 78.43 & \cellcolor{lightorange}81.12 & 81.23 & 68.37 & \cellcolor{lightblue}82.20 &
      49.49 & 50.43 & \cellcolor{lightblue}58.89 & 55.89 & 33.40 & \cellcolor{lightorange}56.16 & 
      49.94 & 53.05 & 55.53 & \cellcolor{lightblue}56.76 & 43.21 & \cellcolor{lightorange}56.30 \\
&amh & 42.45 & 68.83 & \cellcolor{lightorange}72.60 & \cellcolor{lightblue}76.43 & 49.46 & 58.28 &
      29.32 & \cellcolor{lightorange}39.48 & 24.27 & \cellcolor{lightblue}47.09 & 28.19 & 28.43 & 
      41.21 & 26.94 & 42.47 & \cellcolor{lightblue}51.16 & 41.28 & \cellcolor{lightorange}45.75 \\
\midrule 
\multirow{2}{*}{3}
&zho & 83.96 & \cellcolor{lightorange}86.45 & \cellcolor{lightblue}87.08 & 86.44 & 72.49 & 85.83 
    & 57.39 & 47.83 & \cellcolor{lightblue}69.35 & \cellcolor{lightorange}63.12 & 34.05 & 45.17 
    & \cellcolor{lightorange}40.50 & 37.91 & 39.09 & \cellcolor{lightblue}46.10 & 27.72 & 38.87 \\
&mya & 84.10 & 82.44 & \cellcolor{lightorange}84.33 & \cellcolor{lightblue}86.07 & 70.35 & 82.31 &
    \cellcolor{lightorange}42.62 & 34.54 & 25.55 & \cellcolor{lightblue}55.06 & 26.05 & 42.52 & 
    -- & -- & -- & -- & -- & -- \\
\midrule
4 &khm & 47.58 & 47.58 & \cellcolor{lightblue}76.41 & \cellcolor{lightorange}73.70 & 47.58 & 68.38 & 
    23.06 & 50.12 & \cellcolor{lightblue}65.79 & \cellcolor{lightorange}58.60 & 15.89 & 17.20 & 
    14.54 & 29.77 & \cellcolor{lightblue}35.60 & \cellcolor{lightorange}34.28 & 14.57 & 14.26 \\
\midrule
5&swa & 76.54 & 76.02 & 78.17 & \cellcolor{lightblue}79.04 & 75.93 & \cellcolor{lightorange}78.77 
    & 36.39 & 33.20 & \cellcolor{lightblue}43.35 & \cellcolor{lightorange}40.18 & 35.63 & 38.46 
    & \cellcolor{lightorange}55.69 & 54.32 & 56.21 & \cellcolor{lightblue}56.51 & 53.79 & 54.82 \\
\midrule
6&tel & \cellcolor{lightorange}84.86 & 79.63 & \cellcolor{lightblue}88.93 & \cellcolor{lightblue}88.93 & 65.16 & 78.62 
    & 41.16 & 32.36 & \cellcolor{lightblue}43.68 & \cellcolor{lightorange}42.56 & 30.77 & 36.73 
    & 37.87 & 21.55 & \cellcolor{lightblue}39.79 & \cellcolor{lightorange}39.24 & 29.65 & 35.83 \\
\midrule
7&tur & 69.80 & \cellcolor{lightorange}74.47 & 70.94 & \cellcolor{lightblue}75.02 & 65.42 & 73.06  
    & 36.55 & 43.92 & \cellcolor{lightblue}51.03 & \cellcolor{lightorange}48.41 & 30.76 & 31.33 
    & 36.92 & \cellcolor{lightorange}44.83 & 44.48 & \cellcolor{lightblue}44.92 & 36.83 & 44.36 \\    
\bottomrule
\end{tabular}%
}
\vspace{-0.25cm}
\caption{Average macro-F1(\%) scores for \polardetect, \polartype,  and \polardetect~ across languages and multilingual encoders. The best and second performance scores are highlighted in \colorbox{lightblue}{blue} and \colorbox{lightorange}{orange} respectively.}
\label{tab:results_SLMs}
\vspace{-0.55cm}
\end{table*}

\section{Experimentation and Results}
\label{sec:experimentation}

\subsection{Experimental Setup}

To evaluate \datasetname, we conducted baseline experiments on three polarization detection tasks: (1) classifying texts as polarized or not, (2) identifying polarization types, and (3) detecting polarization manifestations. For data splitting, we used 70\% for training, 10\% for validation, and 20\% for testing, as summarized in \Cref{tab:dataset_composition}. 
All experiments were conducted using the EncouRAGe framework proposed by ~\citet{strich2025encourageevaluatingraglocal}, which provides a standardized evaluation protocol for language model benchmarking.
We benchmarked SLMs and LLMs. The list of evaluated models are listed below and Appendix~\ref{sec:SLMs_and_LLMs_Used}:

\begin{itemize}[noitemsep,leftmargin=*]
 \item \textbf{Fine-tuning SLMs:} We fine-tuned six SLMs, including four general-purpose models: mBERT~\citep{devlin2019bert}, XLM-R~\cite{conneau2019unsupervised}, RemBERT~\citep{chung2020rethinkingembeddingcouplingpretrained}, and LaBSE~\citep{feng2022language}. In addition, we evaluated two models build on social media and multilingual training corpus: twitter-roberta-hate~\citep{antypas-camacho-collados-2023-robust}, a RoBERTa-based encoder specialized for hate-speech detection, and AfroXLMR-large-76L~\citep{adelani2023sib200}, an XLM-R variant optimised for African and other low-resource languages.

 \item \textbf{Evaluating LLMs in Zero- and Few-shot Settings:} 
We evaluated large language models in zero- and few-shot settings, including Qwen2.5-7B-Instruct, Qwen-3-8B, LLaMA-3.1-8B-Instruct, Ministral-3-14B-Instruct-2512, Gemma-3-27B-IT, GPT-4.1-Nano, and GPT-OSS-120B. For brevity, we refer to them as Qwen2.5, Qwen3, LLaMA3.1, Mistral3, Gemma3, GPT4.1 and GPT-OSS respectively. The exact models used are stated in the Appendix section~\ref{sec:SLMs_and_LLMs_Used}.
The prompts used for LLM zero-shot and few-shot settings are shown in Appendix~\ref{sec:Prompts_for_LLMs}.
\end{itemize}

\begin{table*}[!htpb]
\centering
\small
\setlength{\tabcolsep}{3pt} 
\renewcommand{\arraystretch}{1}
\resizebox{\textwidth}{!}{%
\begin{tabular}{l|l| c c c c c c c | c c c c c c c | c c c c c c c}
\toprule
\multirow{2}{*}{\#} &
\multirow{2}{*}{\textbf{Lang.}} &
\multicolumn{7}{c}{\polardetect} &
\multicolumn{7}{c}{\polartype} &
\multicolumn{7}{c}{\polarmanifest} \\
\cmidrule(lr){3-9}\cmidrule(lr){10-16}\cmidrule(lr){17-23}
 && \textbf{Qwen2.5} & \textbf{Qwen3} & \textbf{LLaMA3.1} & \textbf{Mistral3} & \textbf{Gemma3} & \textbf{GPT4.1} & \textbf{GPT-OSS} 
& \textbf{Qwen2.5} & \textbf{Qwen3} & \textbf{LLaMA3.1} & \textbf{Mistral3} & \textbf{Gemma3} & \textbf{GPT4.1} & \textbf{GPT-OSS} 
& \textbf{Qwen2.5} & \textbf{Qwen3} & \textbf{LLaMA3.1} & \textbf{Mistral3} & \textbf{Gemma3} & \textbf{GPT4.1} & \textbf{GPT-OSS} \\
\midrule
\multirow{13}{*}{1}
&eng & 71.19 & 69.42 & 69.17 & 70.67 & \cellcolor{lightblue}79.84 & 77.19 & 77.43
    & 72.45 & 66.76 & 69.08 & 73.79 & 74.37 & 69.59 & \cellcolor{lightblue}77.73 
    & 63.33 & 62.38 & 60.62 & 63.91 & 64.18 & 63.39 & \cellcolor{lightblue}68.33 \\
&deu & 60.02 & 58.76 & 61.53 & 61.34 & \cellcolor{lightblue}73.86 & 70.89 & 69.72
    & 66.61 & 61.26 & 64.17 & 67.31 & \cellcolor{lightblue}71.71 & 63.63 & 70.04 
    & 57.74 & 60.10 & 56.82 & 58.71 & 57.93 & 59.41 & \cellcolor{lightblue}65.18 \\
\cmidrule{2-23}
&urd & 40.97 & 48.46 & 58.55 & 51.26 & \cellcolor{lightblue}75.88 & 72.02 & 70.42
    & 42.03 & 44.24 & 49.27 & 45.11 & \cellcolor{lightblue}50.65 & 46.92 & 49.32 
    & 42.20 & 50.32 & 52.14 & \cellcolor{lightblue}54.74 & 53.22 & 48.44 & 52.46 \\
&ben & 65.62 & 59.75 & 68.90 & 69.83 & \cellcolor{lightblue}80.85 & 76.15 & 80.28
    & 68.96 & 66.10 & 65.71 & 71.05 & \cellcolor{lightblue}74.04 & 72.64 & 77.13 
    & 58.47 & 54.54 & 51.23 & 50.47 & 51.89 & 56.99 & \cellcolor{lightblue}60.34 \\
&hin & 43.94 & 36.53 & 48.52 & 45.67 & \cellcolor{lightblue}67.13 & 59.04 & 58.96
    & 65.32 & 54.48 & 59.96 & 68.80 & \cellcolor{lightblue}75.12 & 69.73 & 71.68 
    & 48.09 & 53.62 & 55.74 & \cellcolor{lightblue}60.50 & 58.43 & 53.04 & 58.53 \\
&ori & 53.35 & 50.14 & 49.30 & 42.38 & 65.03 & 58.39 & \cellcolor{lightblue}67.20
    & 65.87 & 64.74 & 65.11 & 48.61 & \cellcolor{lightblue}75.81 & 66.68 & 71.29
    & 55.64 & 55.22 & 54.27 & 49.76 & 59.46 & 57.88 & \cellcolor{lightblue}61.66 \\
&nep & 49.03 & 53.48 & 65.01 & 61.84 & 85.28 & 74.16 & \cellcolor{lightblue}82.68
    & 69.31 & 67.29 & 65.09 & 73.78 & 80.95 & 72.30 & \cellcolor{lightblue}82.29 
    & 62.67 & 67.08 & 61.88 & 70.01 & 73.34 & 69.59 & \cellcolor{lightblue}76.30 \\
&pan & 40.69 & 40.42 & 49.80 & 41.64 & \cellcolor{lightblue}71.95 & 71.67 & 65.92 
    & 56.51 & 54.20 & 56.98 & 54.21 & \cellcolor{lightblue}68.20 & 61.73 & 66.19 
    & 51.10 & 54.90 & 53.20 & 55.22 & 56.73 & 58.48 & \cellcolor{lightblue}61.94 \\
\cmidrule{2-23}
&spa & 66.64 & 68.25 & 70.42 & 70.04 & 71.43 & 73.64 & \cellcolor{lightblue}76.91 
    & 67.69 & 66.07 & 61.82 & 69.05 & 69.81 & 65.93 & \cellcolor{lightblue}72.18 
    & 60.00 & 59.86 & 57.46 & 57.30 & 59.84 & 60.34 & \cellcolor{lightblue}66.36 \\
&ita & 66.24 & 63.20 & 67.52 & 67.36 & 73.72 & 72.08 & \cellcolor{lightblue}74.60
    & 68.48 & 64.68 & 63.59 & 70.19 & 73.27 & 68.47 & \cellcolor{lightblue}74.70 
    & -- & -- & -- & -- & --& -- & --\\
\cmidrule{2-23}
&rus & 73.77 & 71.01 & 71.12 & 72.88 & 70.57 & 72.85 & \cellcolor{lightblue}74.36
    & 70.26 & 66.33 & 57.78 & 73.43 & 69.88 & 68.47 & \cellcolor{lightblue}77.94 
    & -- & -- & -- & -- & --& --& -- \\
&pol & 57.15 & 58.33 & 63.46 & 57.25 & 68.88 & 65.42 & \cellcolor{lightblue}76.94
    & 70.56 & 62.91 & 65.26 & 68.87 & 75.88 & 70.42 & \cellcolor{lightblue}79.46 
    & -- & -- & -- & -- & --& --& -- \\
\cmidrule{2-23}
&fas & 32.17 & 28.44 & 40.10 & 29.71 & 54.79 & 48.71 & \cellcolor{lightblue}51.61
    & 59.53 & 52.96 & 63.23 & 54.93 & \cellcolor{lightblue}71.17 & 60.74 & 66.33 
    & 58.62 & 55.68 & 56.52 & 55.40 & 56.63 & 58.98 & \cellcolor{lightblue}62.93 \\
\midrule 
\multirow{3}{*}{2}
&hau & 51.40 & 49.25 & 55.37 & 50.44 & 46.36 & 52.18 & \cellcolor{lightblue}55.02
    & 55.41 & 53.65 & 50.69 & 54.59 & 54.04 & 56.84 & \cellcolor{lightblue}59.03 
    & \cellcolor{lightblue}51.01 & 47.28 & 43.09 & 43.56 & 43.66 & 48.43 & 48.33 \\
&arb & 65.22 & 58.82 & 70.02 & 65.47 & \cellcolor{lightblue}80.70 & 77.72 & 78.87
    & 63.22 & 62.77 & 62.62 & 66.17 & 71.22 & 64.39 & \cellcolor{lightblue}72.25 
    & 62.74 & 64.57 & 62.64 & 67.04 & 67.97 & 65.02 & \cellcolor{lightblue}73.93 \\
&amh & 25.16 & 26.97 & 39.14 & 23.50 & \cellcolor{lightblue}74.47 & 50.58 & 57.00
    & 52.85 & 51.55 & 61.67 & 47.26 & \cellcolor{lightblue}77.01 & 61.20 & 67.31 
    & 50.78 & 53.99 & 56.51 & 53.43 & 59.77 & 55.98 & \cellcolor{lightblue}61.52 \\
\midrule  
\multirow{2}{*}{3}
&zho & 58.20 & 59.70 & 70.57 & 62.62 & 82.48 & 78.51 & \cellcolor{lightblue}80.52 
    & 77.38 & 73.32 & 68.54 & 75.32 & 80.63 & 74.76 & \cellcolor{lightblue}81.15 
    & 69.99 & 67.72 & 58.16 & 64.35 & 65.82 & 70.44 & \cellcolor{lightblue}77.90 \\
&mya & 40.63 & 48.31 & 51.68 & 37.72 & \cellcolor{lightblue}78.40 & 58.41 & 59.96
    & 52.65 & 55.76 & 58.48 & 49.23 & \cellcolor{lightblue}67.78 & 58.09 & 59.48 
    & -- & -- & -- & -- & -- & --& -- \\
\midrule      
4&khm & 10.56 & 8.94 & 11.04 & 8.65 & \cellcolor{lightblue}13.32 & 11.25 & 12.53 & 
    50.51 & 46.30 & \cellcolor{lightblue}57.32 & 45.32 & 55.21 & 48.71 & 49.43 & 
    47.96 & 50.04 & \cellcolor{lightblue}50.14 & 49.28 & 49.30 & 48.66 & 48.50 \\
\midrule      
5&swa & 40.50 & 46.76 & 57.42 & 52.29 & 65.64 & 62.00 & \cellcolor{lightblue}66.15
    & 61.31 & 57.52 & 61.63 & 62.87 & 69.49 & 66.34 & \cellcolor{lightblue}68.81 
    & 52.94 & 54.04 & 54.63 & 55.11 & 56.89 & 56.22 & \cellcolor{lightblue}60.41 \\
\midrule      
6&tel & 42.99 & 44.23 & 43.02 & 36.82 & 42.97 & \cellcolor{lightblue}45.23 & 43.51
    & 52.81 & 53.88 & 55.82 & 52.24 & \cellcolor{lightblue}57.88 & 52.70 & 50.95 
    & 50.78 & 54.96 & 54.08 & \cellcolor{lightblue}55.68 & 55.29 & 56.59 & 52.13 \\
\midrule   
7&tur & 64.11 & 58.86 & 66.68 & 66.23 & 76.69 & 74.54 & \cellcolor{lightblue}78.97
    & 67.43 & 63.41 & 62.19 & 70.38 & 75.41 & 70.27 & \cellcolor{lightblue}77.49 
    & 60.38 & 61.54 & 60.84 & 62.94 & 65.08 & 62.73 & \cellcolor{lightblue}71.20\\
\bottomrule
\end{tabular}
}
\vspace{-0.25cm}
\caption{F1-Macro resulting from the zero-shot LLM experiments with the \datasetname~dataset. The highest value per language is highlighted in \colorbox{lightblue}{blue}.}
\label{tab:Zero_shot_LLM_results}
\vspace{-0.45cm}
\end{table*}
\subsection{Results and Analysis}



\textbf{SLMs Models:} Table~\ref{tab:results_SLMs} presents the results of six small language models.
Overall, RemBERT and LaBSE show comparable performance across most languages, achieving the best or the second best macro-F1 scores.
RemBERT is designed to balance representation across high-, mid-, and low-resource languages, while LaBSE relies on bilingual sentence-level alignment between English and low-resource languages.
These training strategies enhance their ability to  mid- and low-resource language understanding, which is reflected in their consistent improvements over mBERT and XLM-R, particularly for languages such as Amharic, Odia, Italian.

The twitter-roberta-hate model is finetuned on English Twitter dataset covering emoji, stance, hate speech, and emotion. This training boosts its performance on English polarization detection, polarization types classification and polarization manifestation recognition. AfroXLMR-large-76L is finetuned upon XLM-R on African languages datasets.
As a result, it performs well on African language in our dataset, including Arabic, Hausa, and Swahili.

Polarization  detection is comparatively easier for multilingual BERT-based models, which achieve relatively high macro-F1 scores.
In contrast, recognizing polarization types (politics, gender, racial, religious, and others) is substantially more challenging. Performance drops even further for polarization manifestation recognition (stereotype, vilification, dehumanization, extreme language, lack of empathy, invalidation), where macro-F1 scores decrease markedly. This gap highlights the limitations of current models in capturing fine-grained and implicit polarization manifestations and the importance of \datasetname~ for advancing research on nuanced polarization understanding.

\textbf{LLMs Performance:} The Tables \ref{tab:results_SLMs}, \ref{tab:Zero_shot_LLM_results}, and \ref{tab:Few_show_LLM_results} present an overall picture of model performance in polarization detection, types, and manifestations across the languages, using the Macro-F1 metric as a measure of accuracy. Across all experiment setups, models consistently achieve their highest scores in Polarization Detection, followed by the classification of Polarization Types, while identifying Polarization Manifestations remains the most challenging task for both encoder models and LLMs likely due to the latent nature of how polarization is expressed, requiring semantic understanding that state-of-the-art models find difficult to generalize across diverse cultural contexts.

Table~\ref{tab:results_SLMs} presents the the performance of encoders (mBERT, XLM-R, RemBERT, LaBSE, twitter-roberta-hate, and afro-xlmr-large), demonstrating significant stability across languages, especially in the \polardetect~task. On the other hand, Table~\ref{tab:Zero_shot_LLM_results} presents a more mixed picture through zero-shot LLM experiments.
While high-resource languages like English (eng) and Chinese (zho) maintain high detection scores (reaching 79.84\% and 82.48\% respectively), the models exhibit a significant drop in performance for languages with unique scripts or less resources.
For example, Khmer (khm) detection scores drop to a range between 8.65\% and 13.32\%, indicating that without prior examples, these generative models may not generalize.
Table~\ref{tab:Few_show_LLM_results} demonstrates the transformative impact of few-shot prompting, where providing the LLMs with a few examples significantly improves performance for many languages. For instance, Urdu (urd) has the detection score rise from 40.97\% in the zero-shot setting to 73.81\% with GPT4.1 in the few-shot setting. In this few-shot setting, LLMs also outperform BERT-family encoders in the complex \polarmanifest~task, for example, with Arabic (arb) reaching a peak of 75.49\%. In summary, while BERT-family models remain the most efficient for binary detection, LLMs with few-shot prompting show potential for handling more complex, fine-grained classification tasks when sufficient examples are provided.

Finally, some of the linguistic families exhibit clustered behaviors. For instance, a language cluster emerges among South Asian languages like Bengali (ben), Hindi (hin), Nepali (nep), and Oriya (ori), as these languages exhibit similar performance improvements in the few-shot LLM experiments, compared to zero-shot settings (Table~\ref{tab:Few_show_LLM_results}). Nevertheless, the most dominant predictor of model performance seems to be resource-tier of the language, which often aligns with geographic and economic regions.


\begin{table*}[h]
\centering
\small
\setlength{\tabcolsep}{2pt} 
\resizebox{\textwidth}{!}{%
\begin{tabular}{l|l | c c c c c c c | c c c c c c c | c c c c c c c}
\toprule
\multirow{2}{*}{\#}&
\multirow{2}{*}{\textbf{Lang.}} &
\multicolumn{7}{c}{\polardetect} &
\multicolumn{7}{c}{\polartype} &
\multicolumn{7}{c}{\polarmanifest} \\
\cmidrule(lr){3-9}\cmidrule(lr){10-16}\cmidrule(lr){17-23}
 && \textbf{Qwen2.5} & \textbf{Qwen3} & \textbf{LLaMA3.1} & \textbf{Mistral3} & \textbf{Gemma3} & \textbf{GPT4.1} & \textbf{GPT-OSS} 
 & \textbf{Qwen2.5} & \textbf{Qwen3} & \textbf{LLaMA3.1} & \textbf{Mistral3} & \textbf{Gemma3} & \textbf{GPT4.1} & \textbf{GPT-OSS} 
 & \textbf{Qwen2.5} & \textbf{Qwen3} & \textbf{LLaMA3.1} & \textbf{Mistral3} & \textbf{Gemma3} & \textbf{GPT4.1} & \textbf{GPT-OSS} \\
\midrule
\multirow{13}{*}{1}
&eng & 77.98 & 77.79 & 74.90 & 75.82 & 79.82 & 76.81 & \cellcolor{lightblue}79.74 
    & 72.17 & 73.52 & 70.48 & 77.33 & 76.96 & 74.01 & \cellcolor{lightblue}79.24 
    & 62.37 & 63.40 & 61.56 & 63.41 & 64.35 & 63.77 & \cellcolor{lightblue}69.40 \\
&deu & 65.49 & 65.79 & 65.93 & 64.34 & 71.26 & \cellcolor{lightblue}71.66 & 69.61 
    & 65.03 & 67.75 & 68.37 & 69.58 & \cellcolor{lightblue}72.44 & 68.31 & 69.61 
    & 58.36 & \cellcolor{lightblue}60.76 & 55.97 & 60.59 & 53.72 & 57.59 & 65.03 \\
\cmidrule{2-23}
&urd & 57.89 & 55.42 & 55.48 & 43.96 & 62.62 & \cellcolor{lightblue}73.81 & 66.26 
    & 40.80 & 49.04 & 47.02 & 45.40 & \cellcolor{lightblue}51.85 & 48.14 & 49.39 
    & 43.73 & 57.34 & 54.55 & \cellcolor{lightblue}59.44 & 58.42 & 49.75 & 56.67 \\
&ben & 66.84 & 63.00 & 59.91 & 65.61 & \cellcolor{lightblue}79.64 & 78.89 & 78.32 
    & 70.14 & 72.51 & 71.14 & 73.37 & 75.11 & 71.12 & \cellcolor{lightblue}76.71 
    & 57.01 & 53.29 & 51.41 & 53.12 & 53.72 & 58.90 & \cellcolor{lightblue}60.36 \\
&hin & 52.92 & 43.45 & 41.42 & 52.13 & 58.54 & \cellcolor{lightblue}62.61 & 56.26 
    & 59.85 & 54.86 & 55.54 & 69.68 & 71.31 & 66.72 & \cellcolor{lightblue}70.99 
    & 47.77 & 51.70 & 54.76 & 59.41 & \cellcolor{lightblue}58.65 & 54.29 & 57.49 \\
&ori & 60.85 & 57.23 & 60.81 & 43.60 & 63.62 & 65.71 & \cellcolor{lightblue}66.09 
    & 63.98 & 66.26 & 67.26 & 50.38 & \cellcolor{lightblue}75.68 & 67.56 & 73.38 
    & 54.71 & 56.56 & 53.24 & 48.69 & 59.73 & 57.57 & \cellcolor{lightblue}62.57 \\
&nep & 64.38 & 74.27 & 64.25 & 69.50 & \cellcolor{lightblue}84.64 & 82.44 & 81.20 
    & 68.49 & 73.98 & 71.72 & 77.16 & 81.57 & 74.45 & \cellcolor{lightblue}82.38 
    & 59.81 & 64.29 & 60.89 & 68.35 & 70.57 & 72.03 & \cellcolor{lightblue}76.92 \\
&pan & 51.19 & 60.47 & 56.38 & 52.95 & 69.77 & \cellcolor{lightblue}72.07 & 69.76 
    & 61.04 & 64.38 & 62.86 & 64.38 & 69.43 & 66.76 & \cellcolor{lightblue}70.57 
    & 50.94 & 53.07 & 53.95 & 55.51 & 57.22 & 58.66 & \cellcolor{lightblue}62.82 \\
\cmidrule{2-23}
&spa & 68.01 & 70.43 & 67.35 & 66.91 & 72.15 & 74.05 & \cellcolor{lightblue}74.61 
    & 68.46 & 68.96 & 64.16 & 70.36 & 71.67 & 68.65 & \cellcolor{lightblue}73.39 
    & 59.49 & 58.69 & 58.10 & 59.80 & 60.00 & 62.54 & \cellcolor{lightblue}66.29 \\
&ita & 59.37 & 55.83 & 48.57 & 51.98 & 65.92 & 71.82 & \cellcolor{lightblue}72.02 
    & 60.67 & 63.55 & 63.15 & 64.49 & 70.46 & 68.51 & \cellcolor{lightblue}72.98 
    & -- & -- & -- & -- & --& --& -- \\
\cmidrule{2-23}
&rus & \cellcolor{lightblue}74.96 & 73.70 & 62.74 & 70.59 & 74.32 & 74.04 & 77.45 
    & 69.78 & 70.74 & 63.57 & \cellcolor{lightblue}74.10 & 73.97 & 71.19 & 78.71 
    & -- & -- & -- & -- & --& --& -- \\
&pol & 58.07 & 62.86 & 53.14 & 59.10 & 68.38 & 71.54 & \cellcolor{lightblue}79.17 
    & 70.56 & 66.92 & 69.71 & 71.19 & 76.91 & 72.50 & \cellcolor{lightblue}79.79 
    & -- & -- & -- & -- & --& --& -- \\
\cmidrule{2-23}
&fas & 40.77 & 41.32 & 40.45 & 39.32 & \cellcolor{lightblue}54.58 & 54.22 & 50.53 
    & 57.14 & 59.51 & 66.76 & 59.26 & \cellcolor{lightblue}69.72 & 62.67 & 65.54 
    & 58.08 & 55.41 & 54.25 & 56.35 & 57.53 & 59.24 & \cellcolor{lightblue}62.47 \\
\midrule
\multirow{3}{*}{2}
&hau & 55.01 & 53.92 & 55.60 & 50.47 & 56.24 & 55.08 & \cellcolor{lightblue}59.28 
    & 57.89 & 55.73 & 55.49 & 54.61 & 57.79 & 56.69 & \cellcolor{lightblue}60.64 
    & 50.80 & 49.01 & 47.69 & 46.19 & 45.98 & \cellcolor{lightblue}50.87 & 49.94 \\
&arb & 75.54 & 72.88 & 73.42 & 72.21 & \cellcolor{lightblue}79.58 & 78.37 & 78.40 
    & 64.36 & 65.53 & 66.16 & 70.07 & 71.44 & 70.50 & \cellcolor{lightblue}72.89 
    & 64.89 & 66.55 & 61.96 & 67.07 & 67.6 & 67.58 & \cellcolor{lightblue}75.49 \\
&amh & 38.51 & 35.70 & 37.66 & 28.59 & \cellcolor{lightblue}74.58 & 51.81 & 54.67 
    & 50.41 & 59.07 & 64.44 & 55.74 & \cellcolor{lightblue}76.65 & 59.64 & 66.07 
    & 50.90 & 55.80 & 51.50 & 53.36 & 58.72 & 52.91 & \cellcolor{lightblue}61.37 \\
\midrule
\multirow{2}{*}{3}
&zho & 56.69 & 68.17 & 56.79 & 66.71 & 69.81 & \cellcolor{lightblue}78.22 & 74.57 
    & 71.13 & 78.00 & 72.03 & 76.32 & 76.87 & \cellcolor{lightblue}79.12 & 77.00 
    & 69.97 & 67.39 & 61.25 & 62.78 & 63.98 & 72.01 & \cellcolor{lightblue}76.47 \\
&mya & 62.17 & 71.12 & 60.51 & 50.96 & \cellcolor{lightblue}73.90 & 68.46 & 63.07 
    & 56.23 & 63.35 & 68.11 & 58.24 & 71.63 & \cellcolor{lightblue}65.54 & 63.89 
    & -- & -- & -- & -- & --& --& -- \\
\midrule
4&khm & \cellcolor{lightblue}24.70 & 13.66 & 15.75 & 12.98 & 18.90 & 16.52 & 15.69 
    & 53.99 & 51.21 & \cellcolor{lightblue}62.92 & 51.81 & 57.24 & 51.99 & 51.41 
    & 49.45 & \cellcolor{lightblue}50.55 & 49.55 & 52.19 & 50.47 & 48.27 & 50.11 \\
\midrule
5&swa & 53.67 & 58.43 & 58.75 & 59.20 & \cellcolor{lightblue}67.70 & 64.56 & 66.61 
    & 61.14 & 65.17 & 65.74 & 66.71 & \cellcolor{lightblue}70.80 & 67.59 & 69.81 
    & 56.27 & 54.88 & 53.13 & 53.58 & 56.04 & 57.62 & \cellcolor{lightblue}61.13 \\
\midrule
6&tel & 58.52 & 56.59 & 53.40 & 54.56 & \cellcolor{lightblue}60.02 & 59.74 & 52.15 
    & 52.15 & 55.13 & 56.16 & 56.64 & \cellcolor{lightblue}57.24 & 56.24 & 53.92 
    & 53.79 & 55.79 & 55.41 & 54.61 & \cellcolor{lightblue}59.03 & 56.71 & 52.53 \\
\midrule
7&tur & 72.29 & 68.12 & 66.23 & 68.70 & \cellcolor{lightblue}79.62 & 76.45 & 78.28 
    & 64.02 & 66.43 & 62.74 & 71.08 & 76.03 & 69.09 & \cellcolor{lightblue}77.67 
    & 60.74 & 58.18 & 60.60 & 64.70 & 64.83 & 64.14 & \cellcolor{lightblue}72.87\\
\bottomrule
\end{tabular}
}
\vspace{-0.25cm}
\caption{F1-Macro scores(in \%) from the few-shot LLM experiments with the \datasetname~ dataset. The highest value per language is highlighted in \colorbox{lightblue}{blue}.}
\label{tab:Few_show_LLM_results}
\vspace{-0.35cm}
\end{table*}

\vspace{-0.15cm}
\section{Discussion}

\paragraph{Performance within the same language family:} It is not necessary that models uniformly achieve a high performance across all languages within the same language family.
For example, while a model may perform strongly on Chinese, its performance on Burmese still lags behind, demonstrating that high-resource language performance does not guarantee transferability to related lower-resource languages such as Burmese.

\paragraph{Few-shot vs. Zero-shot:} Few-shot performance is not always superior to zero-shot results.
For larger models, such as Gemma3 and GPT-OSS, the few-shot setting underperforms zero-shot, indicating that large LLMs are inherently capable of recognizing polarized sentences without requiring in-context exemplars.
In contrast, for smaller models, including Qwen2.5, Qwen3, LLaMA3.1, and Mistral3, in-context examples (3-shot prompting) can improve performance, showing that smaller LLMs benefit more from in-context learning.

\paragraph{LLMs vs. SLMs:} SLMs demonstrate improved performance in detecting polarization after fine-tuning, but still struggle to distinguish polarization types and manifestations, reflecting limited semantic knowledge of domain-specific concepts.
Conversely, LLMs generally exhibit a stronger understanding of social-science polarization constructs, enabling better recognition of polarization types and manifestations, yet they remain weaker in direct polarization classification tasks.
This suggests that LLMs possess richer implicit knowledge of social-science constructs, whereas SLMs currently lack comparable semantic grounding for fine-grained polarization distinctions.

\section{Error Analysis: Misclassification Cases}

Based on our error analysis (See detail in Appendix~\ref{sec:error_examples} Table~\ref{sec:error_examples_tab}), we identified a misalignment between the model's classification logic and human judgment of polarization. The model relies on a deterministic, surface-level heuristic: it classifies text as polarized only when it detects explicitly named and opposed groups within the sentence (e.g., "Israeli forces vs. Palestinians").
Conversely, if the text expresses hostility but names only one group or relies on implied opposition, the model defaults to labeling it as non-polarized.
This error stems from models' reliance on textual pattern alone, while human annotators draw upon cultural and contextual knowledge to interpret hostility and implicit group conflict.

\section{Implications}

\textbf{Theoretical Implications:} Our dataset highlights the complexity of online polarization, emphasizing its deep cultural and contextual nature. It reveals the current limits of NLP models in detecting implicit rhetorical tactics, underscoring the need for culturally-aware frameworks.

\textbf{Practical Implications:} The dataset provides a valuable benchmark for developing and evaluating models capable of detecting nuanced forms of online polarization across multiple languages and contexts. It supports the creation of more culturally sensitive and robust tools for the monitoring and mitigation efforts of online discourse.

\textbf{Methodological Implications:} Our multi-label, multi-platform annotation approach underscores the importance of culturally sensitive and detailed labeling strategies. The variability of model performance across languages and contexts indicates a pressing need for methods that integrate cultural signals, multimodal data, and contextual embeddings to improve robustness and reduce performance disparities in social NLP applications.


\section{Conclusion}

In this study, we introduced \datasetname, a comprehensive, multilingual, and multi-event dataset designed to advance the understanding and detection of online polarization across diverse linguistic and cultural contexts. By annotating over 110,000 instances along three critical dimensions, we created a nuanced, fine-grained resource. This dataset captures the complex rhetorical tactics and social dimensions that underpin polarized discourse. Our extensive benchmarking of SOTA SLMs and LLMs reveals that while current models are reasonably effective at binary polarization detection, they face significant challenges in accurately identifying polarization types and rhetorical manifestations, especially in low-resource and culturally nuanced settings. These findings emphasize the deep contextual and implicit nature of online polarization and highlight the limitations of existing NLP approaches. Importantly, our work underscores the critical need for culturally aware, adaptable, and context-sensitive models to effectively monitor and mitigate digital polarization globally. The resources and benchmarks provided herein aim to catalyze future research, fostering the development of more inclusive and robust tools for analyzing social phenomena in multilingual and multicultural online environments.

\section*{Limitations}

While \datasetname{} represents an important step toward multilingual, multicultural, and multievent polarization analysis, several limitations remain. First, annotator understanding - particularly in crowdsourced setups - was sometimes limited, potentially impacting label quality. We mitigated this through strict quality assurance methods, including control questions, pre-study surveys, and ongoing annotator assessment, but some variability in interpretation may persist.

Second, in-house annotation, while yielding higher consistency, sometimes introduced psychological challenges for annotators given the sensitive or hostile nature of polarized content. To address this, we provided detailed instructions and support resources to reduce stress and clarify expectations, but some emotional burden may have remained.

Third, our choice of models is not exhaustive. Although we included several leading multilingual models and both open and closed LLMs. Adding more language-specific models in the future could improve results, especially for monolingual scenarios.

Finally, for some of the languages in our benchmark, the available data size is still limited, which may constrain the generalizability of model training and evaluation for those cases. Future work should expand dataset size and diversity, and explore language- or region-specific model development to better support underrepresented contexts.

\subsection*{Ethics Statement}

This research uses only publicly available, anonymized data and addresses sensitive topics around polarization in diverse cultures. All annotation was conducted by native speakers using culturally appropriate guidelines; annotators were informed of the project’s social good aims, possible distress, and could opt out anytime. Annotators received prompt and fair compensation above local wage standards or per Prolific’s requirements. Despite rigorous protocols, labeling polarization remains subjective; we encourage responsible, ethically grounded use of this resource and discourage misuse.



\bibliography{custom}

\clearpage

\appendix

\section{Language and its language family}
\label{sec:langauge_family}

The~\datasetname{} dataset covers 22 languages from seven linguistic families. The coverage language and their corresponding branches is presented below:

\begin{table}[h]
\scriptsize
\centering
\begin{tabular}{l|lcrr}
\toprule
{\#}&\textbf{Lang.} & \textbf{ISO-639} & \textbf{Language Family} & \textbf{Sub-branch} \\
\midrule
\multirow{13}{*}{1}
&English & eng & Indo-European & Germanic \\
&German & deu & Indo-European & Germanic \\
\cmidrule{2-5}
&Urdu & urd & Indo-European & Indo-Aryan \\
&Bengali & ben & Indo-European & Indo-Aryan \\
&Hindi & hin & Indo-European & Indo-Aryan \\
&Odia & ori & Indo-European & Indo-Aryan \\
&Nepali & nep & Indo-European & Indo-Aryan \\
&Punjabi & pan & Indo-European & Indo-Aryan \\
\cmidrule{2-5}
&Spanish & spa & Indo-European & Romance \\
&Italian & ita & Indo-European & Romance \\
\cmidrule{2-5}
&Russian & rus & Indo-European & Slavic \\
&Polish & pol & Indo-European & Slavic \\
\cmidrule{2-5}
&Persian & fas & Indo-European & Iranian \\
\midrule
\multirow{3}{*}{2}
&Hausa & hau & Afro-Asiatic & Chadic \\
&Arabic & arb & Afro-Asiatic & Semitic \\
&Amharic & amh & Afro-Asiatic & Semitic \\
\midrule
\multirow{2}{*}{3}
&Chinese & zho & Sino-Tibetan & Sinitic \\
&Burmese & mya & Sino-Tibetan & Tibeto-Burman \\
\midrule
4&Khmer & khm & Austroasiatic & Mon-Khmer\\
\midrule
5&Swahili & swa & Niger–Congo & Bantu \\
\midrule
6&Telugu & tel & Dravidian & Dravidian \\
\midrule
7&Turkish & tur & Turkic & Turkic \\
\bottomrule
\end{tabular}
\caption{Language covered and its language families}
\label{tab:language_family}
\end{table}

\section{Data Statistics}
\label{app:data_stats}


Table~\ref{tab:data_sources} summarizes the distribution of instances across data sources in the \datasetname{} dataset. The data primarily originate from major social media platforms, with Twitter contributing over half of the instances. Additional data are drawn from news websites, online forums, and other social platforms to ensure coverage of diverse discourse contexts. A smaller portion of the data comes from existing datasets that were re-annotated to align with our polarization framework.

\begin{table}[!htpb] 
\scriptsize
\centering
\begin{tabular}{lrr} 
\toprule 
\textbf{Source} & \textbf{No. of Instances} & \textbf{\%} \\ 
\midrule 
X (Twitter) & 58,984 & 53.3\% \\ 
News/websites & 15,208 & 13.7\% \\ 
Bluesky & 8,099 & 7.3\% \\ 
YouTube & 7,159 & 6.5\% \\ 
Reddit & 5,705 & 5.2\% \\ 
Existing Dataset & 4,794 & 4.3\% \\ 
Weibo & 4,550 & 4.1\% \\ 
Facebook & 2,928 & 2.6\% \\ 
Zhihu & 1,088 & 1.0\% \\ 
Threads & 852 & 0.8\% \\ 
Tieba & 783 & 0.7\% \\ 
Wikipedia & 500 & 0.5\% \\ 
\midrule 
\textbf{Total} & \textbf{110,650} & \textbf{100\%} \\ 
\bottomrule 
\end{tabular} 
\caption{Data sources of the \datasetname{} dataset.} 
\label{tab:data_sources} 
\end{table} 

\subsection{Dataset Composition by Language}
\label{app:dataset_composition}

Table~\ref{tab:dataset_composition} presents a language-wise overview of the \datasetname{} dataset, including data sources, targeted events or topics, and inter-annotator agreement. For each language, data collection was tailored to platform availability and sociopolitical relevance, resulting in variation in event focus and discourse type. Inter-annotator agreement is reported primarily using Fleiss’s kappa, with alternative reliability measures noted where applicable.

\subsection{Polarization Statistics}
\label{app:polarization_stats}

\begin{figure*}[h]
    \centering
    \includegraphics[width=0.95\linewidth]{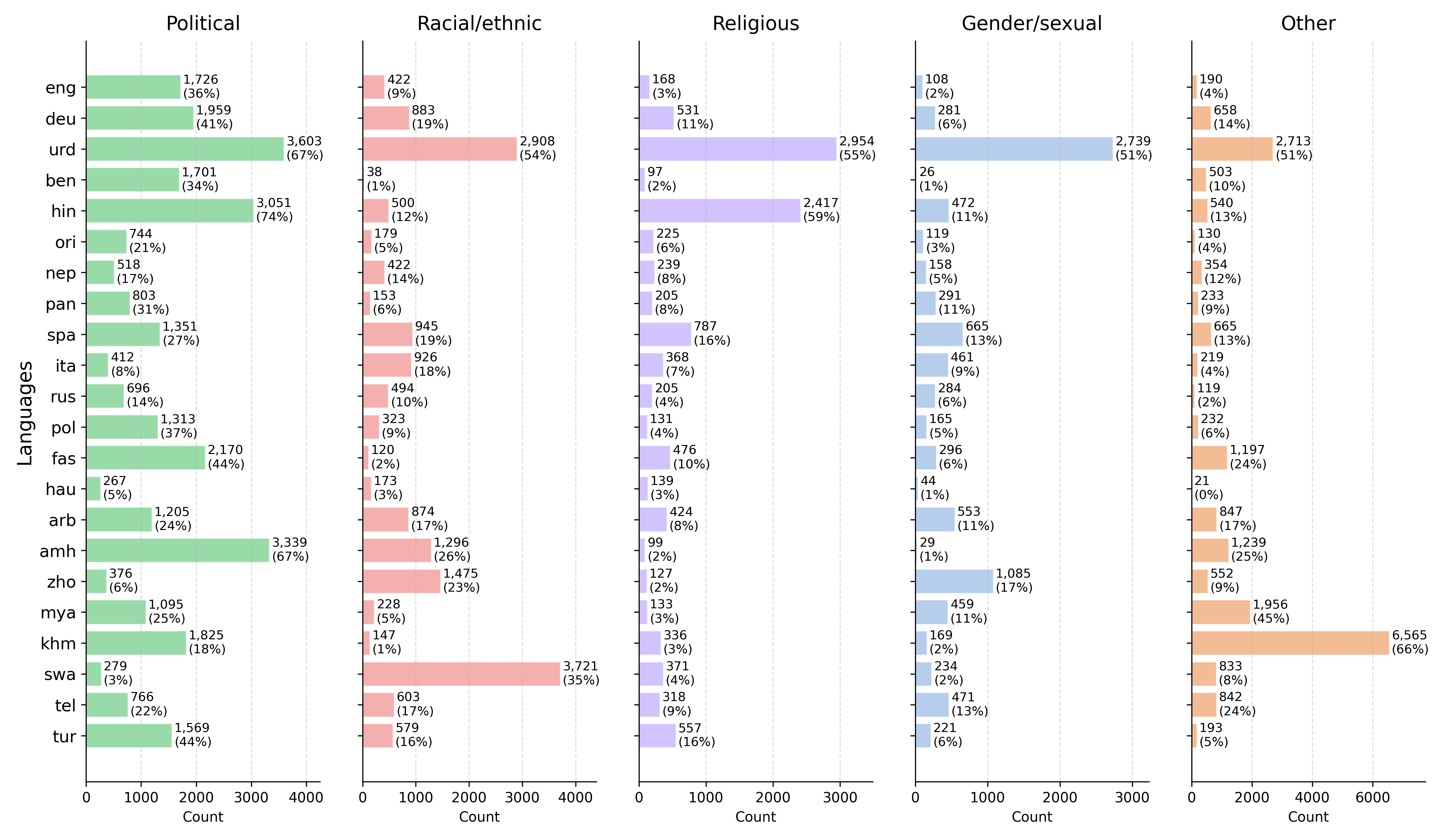}
    \caption{\polartype~ by  Languages, for each language, the numeric value shows the count of instances assigned to a given polarization type, while the percentage reflects its share of the total annotated instances for that language.}
    \label{fig:polarty_type}
\end{figure*}

\begin{figure*}[h]
    \centering
    \includegraphics[width=0.95\linewidth]{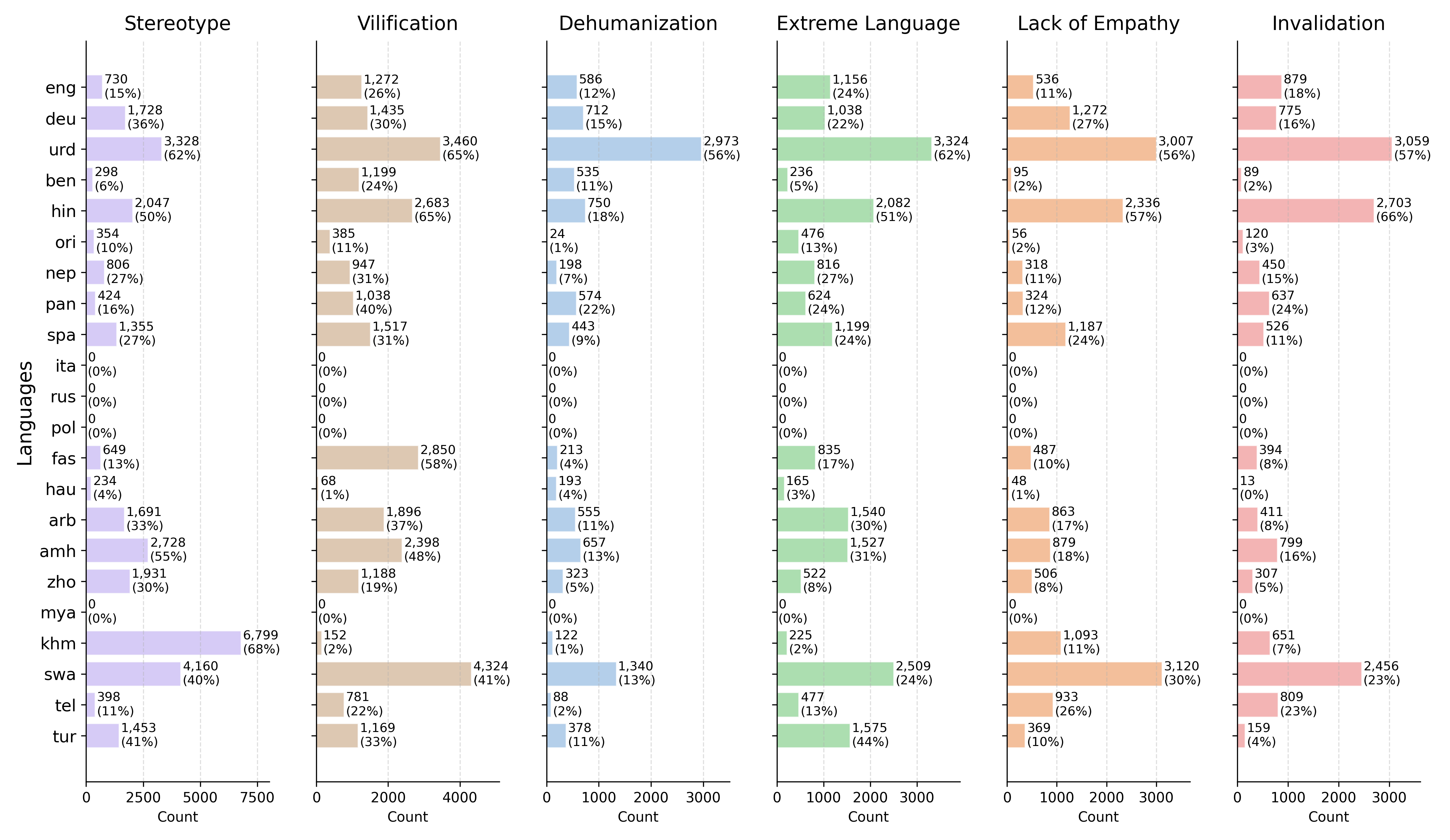}
    \caption{\polarmanifest-by Languages. For each language, the numeric value shows the count of instances assigned to a given polarization manifestation, while the percentage reflects its share of the total annotated instances for that language.} 

    \label{fig:polarity_manifestion}
\end{figure*}

Figure~\ref{fig:polarty_type} shows the distribution of polarization types, and Figure~\ref{fig:polarity_manifestion} illustrates the distribution of polarization manifestations across languages.

\begin{table*}[!ht]
\centering
\scriptsize
\setlength{\tabcolsep}{4pt}
\renewcommand{\arraystretch}{1}
\begin{tabular}{lll}
\hline
\textbf{Language} & \textbf{Data Source(s)}                                                                                                          & \textbf{Events/Topics Focused}                                                                                                                                          \\ \hline
Amharic (amh)     & Facebook, X (Twitter)                                                                                                            & The Tigray War                                                                                                                                                          \\ \hline
Arabic (arb)      & \begin{tabular}[c]{@{}l@{}}Existing datasets, Facebook, \\ News, Reddit, Threads,X (Twitter)\end{tabular}                        & Social issues regading politics and religion                                                                                                                            \\ \hline
Bengali (ben)     & YouTube comments                                                                                                                 & Social discourse, generic contemporary topics                                                                                                                           \\ \hline
Burmese (mya)     & Existing datasets \cite{10613636}, Wikipedia                                                                                                     & \begin{tabular}[c]{@{}l@{}}Social issues regading politics, ethnicity \\ and popular culture\end{tabular}                                                               \\ \hline
Chinese (zho)     & Tieba, Weibo, Zhihu                                                                                                              & \begin{tabular}[c]{@{}l@{}}Social issues regarding racism, \\ sexuality/gender and religious discrimination\end{tabular}                                                \\ \hline
English (eng)     & Bluesky, Local news, X (Twitter)                                                                                                 & US elections and international conflicts                                                                                                                                \\ \hline
German (deu)      & Bluesky, Reddit, X (Twitter)                                                                                                     & COVID-19, and contemporary social issues                                                                                                                                \\ \hline
Hausa (hau)       & Facebook, X (Twitter)                                                                                                            & \begin{tabular}[c]{@{}l@{}}Social issues regading politics, ethnicity \\ and religion\end{tabular}                                                                      \\ \hline
Hindi (hin)       & Bluesky, Reddit, X (Twitter)                                                                                                     & \begin{tabular}[c]{@{}l@{}}Social issues regarding politics, religion,\\ and caste\end{tabular}                                                                         \\ \hline
Italian (ita)     & YouTube, X (Twitter)                                                                                                             & \begin{tabular}[c]{@{}l@{}}Pride parade, immigration crisis\\ , crime news, Italian justice reform\end{tabular}                                                         \\ \hline
Khmer (khm)       & \begin{tabular}[c]{@{}l@{}}Facebook, Specialised websites, \\ Wikipedia, YouTube, Local news\end{tabular}                        & COVID-19, and contemporary social issues                                                                                                                                \\ \hline
Nepali (nep)      & Facebook, Local news, X (Twitter)                                                                                                & Social discourse, generic contemporary topics                                                                                                                           \\ \hline
Odia (ori)        & Bluesky, Local news, X (Twitter)                                                                                                 & Social discourse, generic contemporary topics                                                                                                                           \\ \hline
Persian (fas)     & Bluesky, X (Twitter)                                                                                                             & Social discourse, generic contemporary topics                                                                                                                           \\ \hline
Polish (pol)      & Bluesky, Existing dataset \cite{kołos2024banpl}                                                                                                       & COVID-19, and contemporary social issues                                                                                                                                \\ \hline
Punjabi (pan)     & \begin{tabular}[c]{@{}l@{}}Existing dataset (pnbTenTen) \\
\url{https://www.sketchengine.eu/pnbtenten-western-punjabi-corpus}
, \\ Youtube Comments, \\ X (Twitter)\end{tabular}                                      & \begin{tabular}[c]{@{}l@{}}Social issues regarding politics, religion,\\ and caste\end{tabular}                                                                         \\ \hline
Russian (rus)     & \begin{tabular}[c]{@{}l@{}}Bluesky, X (Twitter), \\ COVID-19 chatter dataset \cite{banda2021large}\end{tabular} & Social discourse, generic contemporary topics                                                                                                                           \\ \hline
Spanish (spa)     & Bluesky, X (Twitter)                                                                                                             & \begin{tabular}[c]{@{}l@{}}2010's inmigration movement, \\ "Salvemos las dos vidas" movement, \\ social issues regarding politics \\ and gender inequality\end{tabular} \\ \hline
Swahili (swa)     & X (Twitter)                                                                                                                      & Kenyan elections                                                                                                                                                        \\ \hline
Telugu (tel)      & Facebook, Reddit, X (Twitter)                                                                                                    & Social discourse, generic contemporary topics                                                                                                                           \\ \hline
Turkish (tur)     & X (Twitter), Existing dataset \cite{coltekin-2020-corpus}                                                                                                   & Social discourse, generic contemporary topics                                                                                                                           \\ \hline
Urdu (urd)        & X (Twitter)                                                                                                                      & Social discourse, generic contemporary topics                                                                                                                           \\ \hline
\end{tabular}
\caption{Summary of dataset composition regarding data sources and events or topics focused.}
\label{tab:dataset_composition}
\end{table*}

\section{SLMs and LLMs Used}
\label{sec:SLMs_and_LLMs_Used}

\subsection{Multilingual Encoders}
\begin{itemize}
  \item \url{https://huggingface.co/google-bert/bert-base-multilingual-cased}
  \item \url{https://huggingface.co/FacebookAI/xlm-roberta-base}
  \item \url{https://huggingface.co/google/rembert}
  \item \url{https://huggingface.co/sentence-transformers/LaBSE}
  \item \url{https://huggingface.co/cardiffnlp/twitter-roberta-base-hate}
  \item \url{https://huggingface.co/Davlan/afro-xlmr-large-76L}
\end{itemize}

\subsection{LLMs}
\begin{itemize}
  \item \url{https://huggingface.co/Qwen/Qwen2.5-7B-Instruct}
  \item \url{https://huggingface.co/Qwen/Qwen3-8B}
  \item \url{https://huggingface.co/meta-llama/Llama-3.1-8B-Instruct}
  \item \url{https://huggingface.co/mistralai/Ministral-3-14B-Instruct-2512}
  \item \url{https://huggingface.co/google/gemma-3-27b-it}
  \item \url{https://huggingface.co/openai/gpt-oss-120b}
\end{itemize}

\section{Experiment Settings}



\subsection{Experiments settings}

For SLMs, we performed language-specific fine-tuning for 3 epochs using a learning rate of 2e-4.

For LLMs in the few-shot setting , we use three shots. ALL prompts are written in English, while the in-context examples are provided in the target language. The full prompts is reported in Appendix~\ref{sec:Prompts_for_LLMs}.

All experiments were conducted on local GPU and implemented with EncouRAGe~\citep{strich2025encourageevaluatingraglocal} framework to execute experiments and measure results.

\onecolumn 
\section{Annotation Guidelines}
\label{sec:Annotation_Guidelines}

\begin{tcolorbox}[
  colback=white,
  colframe=black,
  boxrule=0.8pt,
  arc=3pt,
  left=4pt, right=4pt, top=4pt, bottom=4pt,
  fonttitle=\bfseries,
]

\begin{center}
\LARGE \textbf{Annotation Guidelines}
\end{center}
\vspace{1em}
To minimize potential misunderstandings, we offer annotation guidelines customized for each event, accompanied by event-specific examples. These guidelines are in the languages of the target datsets. The English version is provided below.
\vspace{1em}
This guideline aims to assess whether social media messages reflect \textbf{attitude polarization} and to categorize the various types and manifestations of this polarization. The dataset consists of content sourced from platforms such as Facebook and Twitter.

\vspace{1em}
\noindent
\textcolor{red}{\textbf{Warning to annotators:}} You may encounter polarized and hateful content during this task. Please take breaks as needed and prioritize your mental well-being.

\vspace{1em}
\noindent
If you choose to participate, please read the following sections carefully. To exit the study at any time, click ``\textbf{Cancel}'' or ``\textbf{Quit}'' and provide feedback if possible.

\section*{Task 1: Polarization Classification (Yes/No)}

Polarization refers to increasingly extreme, divided beliefs or behaviors between opposing groups. \textbf{Attitude polarization} includes:

\begin{itemize}
  \item Negative attitudes toward out-groups
  \item Blind support for in-groups
  \item Stereotyping, vilification, dehumanization, or intolerance
\end{itemize}

\noindent
Texts should be labeled:
\begin{itemize}
  \item \textbf{Yes} – if the message clearly reflects attitude polarization
  \item \textbf{No} – if it does not show any polarization indicators
\end{itemize}

\textbf{Note:} Always consider the overall context and meaning, not just individual words.

\section*{Task 2: Type of Polarization (if ``Yes'' in Task 1)}

If the message is polarized, select all applicable types:

\begin{enumerate}[label=2.\arabic*]
  \item \textbf{Political/Ideological Polarization} – Conflict between political parties or ideologies
  \item \textbf{Racial/Ethnic Polarization} – Division based on race or ethnicity
  \item \textbf{Religious Polarization} – Conflict based on religious beliefs
  \item \textbf{Gender/Sexual Polarization} – Discrimination based on gender or sexual orientation 
  \item \textbf{Other} – e.g., based on economic class, technology, media, etc.
\end{enumerate}

\textbf{Note:} Select all relevant options if the text contains more than one of the above.

\section*{Task 3: Manifestations of Polarization (if ``Yes'' in Task 1)}

Classify the tactics used to express polarization in the message. A text is considered polarizing if it includes one or more of the following:

\begin{enumerate}[label=3.\arabic*]
  \item \textbf{Stereotyping} – Generalizing traits to all group members  
    \textit{Example: ``Women are weak.''}
    
  \item \textbf{Vilification} – Defaming or demonizing a group  
    \textit{Example: ``Migrants are traitors.''}
    
  \item \textbf{Dehumanization} – Stripping away human qualities  
    \textit{Example: ``They are cockroaches.''}
    
  \item \textbf{Extreme Language or Absolutism} – Using words like ``always'', ``never'', or ``worst''.  
    \textit{Example: ``We can never trust them.''}
    
  \item \textbf{Lack of Empathy or Understanding} – Dismissing others’ perspectives  
    \textit{Example: ``Wearing the hijab is extremist.''}
    
  \item \textbf{Invalidation} – Denying a group’s identity or existence  
    \textit{Example: ``There is no nation called Palestine.''}
\end{enumerate}
\textbf{Note:} Select all relevant options if the text contains more than one of the above.
\end{tcolorbox}

\section{Prompts for Text Classification}
\label{sec:Prompts_for_LLMs}

\subsection{Prompts for polarization detection}
\label{sec:Prompt for Text Classification'}
\begin{tcolorbox}[
  colback=white,
  colframe=black,
  boxrule=0.8pt,
  arc=3pt,
  left=4pt, right=4pt, top=4pt, bottom=4pt,
  fonttitle=\bfseries,
]

\textbf{TASK}: Determine whether the following \{\{lang\_full\}\} text is polarized or not polarized.
\vspace{1em}

\textbf{DEFINITION}:
Polarization refers to the process or phenomenon in which opinions, beliefs, or behaviors become more extreme or divided, leading to a greater distance or conflict between differing groups.
Attitude polarization is the negative attitude that individuals or groups display towards individuals and groups outside their group while also showing blind support and solidarity towards people within their group.\\

Polarization denotes stereotyping, vilification, dehumanization, deindividuation, or intolerance of other people’s views, beliefs, and identities.
Speeches and articles that are shared on social media that incite division, groupism, hatred, conflict, and intolerance are classified as containing polarization.
Only texts that clearly reflect attitude polarization should be classified as such.\\

If a text includes one or more of the specified characteristics, it is classified as polarized. Conversely, social media texts that do not display any of these characteristics are classified as non-polarized.
Always consider the context and the overall meaning of the text, not just individual words or phrases.
\vspace{1em}

\textbf{INSTRUCTIONS}:\\
- Analyze the text carefully.\\
- Classify the polarization of the given text. Use the given definition of polarization.\\
- Provide a short reason explaining your decision. The reason should be English.\\
- Assign a binary label: If the text is polarized, polarization is set to 1. If the text is not polarized, polarization is set to 0.\\
- Do not include any text or formatting outside the JSON.\\
\vspace{1em}

\textbf{OUTPUT FORMAT}:

\begin{verbatim}
{"reason": "the reason of the classification decision", "polarization": 0 or 1}
\end{verbatim}
\vspace{1em}
\textbf{Examples (if applicable):}

\begin{verbatim}
{% if examples %}
{{examples}}
{% endif %}
\end{verbatim}
\vspace{1em}
\textbf{Input Text:}
\begin{verbatim}
{{user_prompt}}
\end{verbatim}
\end{tcolorbox}

\newpage
\subsection{Prompt for polarization type}
\label{sec:Prompt for type Classification}

\begin{tcolorbox}[
  colback=white,
  colframe=black,
  boxrule=0.8pt,
  arc=3pt,
  left=4pt, right=4pt, top=4pt, bottom=4pt,
  fonttitle=\bfseries,
]
\textbf{TASK}: Determine whether the following \{\{lang\_full\}\} text is polarized or not polarized and which type of polarization.

\textbf{DEFINITION}:
Polarization refers to the process or phenomenon in which opinions, beliefs, or behaviors become more extreme or divided, leading to a greater distance or conflict between differing groups.
Polarization denotes stereotyping, vilification, dehumanization, deindividuation, or intolerance of other people’s views, beliefs, and identities.
Only texts that clearly reflect attitude polarization should be classified as such.
If a text includes one or more of the specified characteristics, it is classified as polarized. Conversely, social media texts that do not display any of these characteristics are classified as non-polarized.
Always consider the context and the overall meaning of the text, not just individual words or phrases.

\textbf{TYPES OF POLARIZATION}:\\
- Political/ideological polarization: Political polarization refers to political beliefs and affiliations becoming more extreme. \\
- Racial or ethnic polarization: This type of polarization focuses on ethnic identity or racial origin and incites division, intolerance, and conflict between ethnic groups or races. \\
- Religious polarization: This type of polarization focuses on religious identity and incites division, intolerance, and conflict between religious followers.\\
- Gender/Sexual polarization: This type of polarization refers to the exclusion, discrimination, and marginalization of individuals based on their gender and sexual orientations within society, often leading to heightened tensions, misunderstandings, or conflicts.\\
- Other: polarization texts targeting other groups/identities such as economy, technology, media, polarization, etc.\\

\textbf{INSTRUCTIONS}:\\
- Analyze the text carefully, and classify the polarization of the given text and if the text is polarized the types of polarization.\\
- Use the given definition of polarization and their types. Provide a short reason explaining your decision. The reason should be English.\\
- Assign a binary label for polarization: If the text is polarized, polarization is set to 1. If the text is not polarized, polarization is set to 0.\\
- Classify the text for the different types of polarization.\\
- The answers need to be in the following order: [political/ideological, racial/ethnic, religious, gender/sexual, other].\\
- The text can contain more than one type of polarization. If the text is not polarized set all polarization types to 0. Do not include any text or formatting outside the JSON.\\

\textbf{OUTPUT FORMAT}:
\begin{verbatim}
{   "reason": "the reason of the classification decision",
    "polarization": 0 or 1, 
    "polarization Types": [0/1, 0/1, 0/1, 0/1, 0/1]
}
{% if examples %}
**EXAMPLES**:
{{examples}}
{% endif %}
**TEXT**:
{{user_prompt}}

\end{verbatim}
\end{tcolorbox}

\subsection{Prompt for polarization manifestation}
\label{sec:Prompt_for_manifestation_Classification}
\begin{tcolorbox}[
  colback=white,
  colframe=black,
  boxrule=0.8pt,
  arc=3pt,
  left=4pt, right=4pt, top=4pt, bottom=4pt,
  fonttitle=\bfseries,
]

\textbf{TASK}: Determine the manifestation type(s) of polarization expressed in the following \{\{lang\_full\}\} text.\\
   
\textbf{LABEL SET}:
["stereotype", "vilification", "dehumanization", "extreme\_language", "lack\_of\_empathy", "invalidation"] \\

\textbf{DEFINITIONS}:\\
- stereotype: Generalizes traits to all members of a group, ignoring individual differences.\\
- vilification: Defames or demonizes a group or person through exaggerated or biased framing.\\
- dehumanization: Denies humanity by comparing people to animals, machines, or objects.\\
- extreme\_language: Uses absolute or hyperbolic language (e.g., “always”, “never”, “us vs. them”).\\
- lack\_of\_empathy: Shows indifference or refusal to understand others’ perspectives.\\
- invalidation: Denies or rejects the identity, legitimacy, or existence of a group or individual.\\

\textbf{INSTRUCTIONS}:\\
- Select all labels from the list that apply (one or more).\\
- If the text is not polarized, set labels = [0].\\
- Provide a short reasoning sentence explaining your decision.\\
- Do not include any text or formatting outside the JSON.\\

\textbf{OUTPUT FORMAT}:
\begin{verbatim}
{
    "reason": "the reason of the classification decision",
    "polarization": 0 or 1, 
    "polarization Types": [0/1, 0/1, 0/1, 0/1, 0/1, 0/1]
}

{% if examples %}
**EXAMPLES**:
{{examples}}
{% end if %}

**TEXT**:
{{user_prompt}}
\end{verbatim}
\end{tcolorbox}

\newpage
\section{Error examples}
\label{sec:error_examples}
\scriptsize
\centering
\begin{table}[ht]

\begin{tabular}{p{\dimexpr0.4\textwidth-2\tabcolsep}| 
                 p{\dimexpr0.1\textwidth-2\tabcolsep}|
                 p{\dimexpr0.1\textwidth-2\tabcolsep}|
                 p{\dimexpr0.4\textwidth-2\tabcolsep}}
\toprule
\textbf{Text} & \textbf{Gold}& \textbf{Type} & \textbf{Reason given by model} \\ 
\midrule
The Cybersecurity Infrastructure Security Agency has frozen its election security   work is reviewing everything it has done to help state local officials secure   elections for the past 8 years. ...The move accommodates Ts false claim of   election fraud. @URL & 0 & FP & The text criticizes a political figure's claim as false and portrays a government agency as accommodating   that claim, expressing a negative attitude toward the out-group and  contributing to partisan division." \\ \midrule
Four key facts Hezbollah is not an Islamic sect or denomination. It is a terrorist org founded in 1985. Hezbollah is an Iranian Shia Islamist proxy based in Lebanon. Not a Lebanese resistance movement. & 0 & FN & The text vilifies Hezbollah by labeling it a terrorist organization and dismissing its claimed legitimacy, showing hostile stereotyping toward the group, which meets the criteria for attitude polarization. \\ \midrule
this new optic the President use that term lightly posted on X of him as a Kingis he serious? Are we no longer a Democracy? We have an election wearly voting starting tomorrow. Im running. Did King Donny stop elections? I appreciate your courage. & 1 & FN & The text merely questions the President's rhetoric and expresses concern about democratic processes without demeaning, vilifying, or stereotyping any group, so it does not show attitude polarization. \\ \midrule
Black Officers Awarded \$16M After White Colleague's Racial Comments & 1 & FN & The headline reports an incident involving race but does not contain demeaning language, vilification, or calls for hostility toward any group, so   it is not polarized. \\ \midrule
And thats not ethnic cleansing how ? & 1 & FN & The brief question does not contain explicit negative or vilifying language toward any group, nor does it promote division or intolerance, so it lacks clear attitude polarization. \\
\bottomrule
\end{tabular}
\caption{Example texts, the human annotated gold labels, the error type (FN=False Negative, FP=False Positive) and the Reason given by the model}
\label{sec:error_examples_tab}
\end{table}

\newpage
\section{Dataset Samples}

The dataset samples are available at the following link:  
\url{https://anonymous.4open.science/r/trial-data-05C6/README.md}.

\label{sec:dataset samples}
\small
\renewcommand{\arraystretch}{1}
\begin{center}
\begin{longtable}{p{9cm} p{1cm} p{2cm}}
\toprule
\textbf{Text} & \textbf{Types} & \textbf{Manifestation} \\
\midrule
\endfirsthead
\toprule
\midrule
\endhead

"Denazification" as an excuse to explain Russia’s deployment of troops to Ukraine, alleging that Hitler was of Jewish ancestry & P & V, EL \\ \midrule
"Heck, I voted for Biden three times." Signed, Liberal Asshats in Swing States & P & EL, S, D \\ \midrule
Satellite images confirm civilian massacres in towns during Russian military occupation & P & V \\ \midrule
"Sufficient evidence of voter fraud?" Where? marjorieisaninsurrectionist & P & EL, I \\ \midrule
"They" are now the deep state. & P & S, I, D \\ \midrule
15000 dead children claimed by Hamas who uses those children as human shields. Yes yes we covered that already. Now you’re just talking in circles & P,Ra & V, EL, I, LE, D \\ \midrule
5 Black Officers Awarded \$16M After White Colleague’s Racial Comments & R & V \\ \midrule
A Federal Investigation needs to be launched against TWITTER Fraud and Election Interference & P & EL, I \\ \midrule
A Jewish state was created because no one else wanted them. This is why. Zionism and Naziism are the same disease. & P+Re+Ra & V, EL, S, I, LE, D \\ \midrule
A lot of lying by the radical left. & P & V, S, I, D \\ \midrule
A rational and just society probably wouldn’t allow school shootings to be part of their National identity & P & EL, S, I, D \\ \midrule
A relationship between the Apartheid state and a tiny tyrant state. Free Palestine & P & V, I, LE, D \\ \midrule
A small price to pay for Ukrainian sovereignty and our green future. Stop whining. & P & I, LE \\ \midrule
A traitor today, a traitor tomorrow, a traitor always! & P & V \\ \midrule
A valid protest of a rigged election, that ended up having some sort of kerfuffle. & P & V, I \\ \midrule
A very moral army. IDF Gaza Israel & P & V, LE \\
\bottomrule
\caption{Examples of polarized statements with annotated types and manifestations.
\textbf{Types}: \textbf{P}: Political, \textbf{Re}: Religious, \textbf{Ra}:Racial;
\textbf{Manifestations}: \textbf{V}: Vilification, \textbf{EL}: Extreme Language, \textbf{S}: Stereotype, \textbf{I}: Invalidation, \textbf{LE}: Lack of Empathy, \textbf{D}: Dehumanization.} \\
\label{tab:sample_data}
\end{longtable}
\end{center}

\begin{table}[ht]
\centering
\small
\renewcommand{\arraystretch}{1}
\begin{tabular}{p{10cm} p{1.5cm} p{1.5cm}}
\toprule
\textbf{Text} & \textbf{Lang} & \textbf{Polarized} \\
\midrule
Gibt sogar recht viele Ex-Muslime. Alles sehr vernünftige Leute. & DE & TRUE \\
There are actually quite a few ex-Muslims. All very reasonable people. & EN &  \\
\hline
Kriegstreiber! Auch ihr, die Propagandapresse!!! & DE & TRUE \\
Warmongers! You too, the propaganda press!!! & EN &  \\
\hline
Steuerzahler und Juden & DE & TRUE \\
Taxpayers and Jews & EN &  \\
\hline
Ja die Türken leiden sehr stark unter ihrem links grün versifften Erdogan. Muss man wissen. :D & DE & TRUE \\
Yes, the Turks are suffering greatly under their left-green filthy Erdogan. Good to know. :D & EN &  \\
\hline
Die Klima Armee Fraktion wird kommen. & DE & TRUE \\
The Climate Army faction will come. & EN &  \\
\hline

Denk ich nicht, nein. Migration ist deren Kernthema. & DE & TRUE \\
I don't think so, no. Migration is their core issue. & EN &  \\
\hline
"Es gibt bereits vereinzelte Einsätze von Bodentruppen in Gaza. @URL" & DE & FALSE \\
"There are already isolated deployments of ground troops in Gaza. @URL" & EN &  \\
\hline
Lass dich nicht aufhalten. Es gibt eine ukrainische Fremdenlegion. Da kannst du dich melden. & DE & FALSE \\
Don't let yourself be stopped. There is a Ukrainian foreign legion. You can sign up there. & EN &  \\
\hline
Klingt für mich nach Long COVID & DE & FALSE \\
Sounds like long COVID to me. & EN &  \\
\hline
Wenn ich schon wieder dieses Woke lese. Der AfD ist sogar unser Grundgesetz zu Woke. Hört auf das Wort so inflationär zu gebrauchen. & DE & FALSE \\
When I read this "woke" again. Even our constitution is too woke for the AfD. Stop using the word so inflationarily. & EN &  \\
\hline

CO2 ist gut für Pflanzen & DE & FALSE \\
CO2 is good for plants. & EN &  \\

\hline
\end{tabular}
\caption{German samples with labels and English translations}
\label{tab:SampleGerman}
\end{table}

\begin{table}[ht]
\centering
\small
\renewcommand{\arraystretch}{1}
\begin{tabular}{p{10cm} p{1.5cm} p{1.5cm}}
\toprule
\textbf{Text} & \textbf{Lang} & \textbf{Polarized} \\
\midrule
y vos un random judio que comenta pelotudeces para poder figurar en algun lado & SPA & TRUE \\
And you, a random Jew who comments nonsense just to get noticed somewhere. & EN &  \\
\hline
quiero comer indio viejo & SPA & TRUE \\
I want to eat old Indian. & EN &  \\
\hline
es que a dia de hoy parece que la diferencia es nazis con aborto y nazis sin aborto & SPA & TRUE \\
Nowadays it seems that the difference is Nazis with abortion and Nazis without abortion. & EN &  \\
\hline
orgulloso de sacarle la conchetumare a un pastor evangelico por meterse en weas que no debia. & SPA & TRUE \\
Proud to beat the hell out of an evangelical pastor for meddling in things he shouldn't. & EN &  \\
\hline
el tribunal superior del partido corrupto & SPA & TRUE \\
The supreme court of the corrupt party. & EN &  \\
\hline
claudia es judia, que no manche & SPA & FALSE \\
Claudia is Jewish, don't stain. & EN &  \\
\hline
la escuela feminista de pintura & SPA & FALSE \\
The feminist school of painting. & EN &  \\
\hline
la cancion que me representa ahora que estoy deportado de los estados unidos. & SPA & FALSE \\
The song that represents me now that I am deported from the United States. & EN &  \\
\hline
aunque, siendo justos, esa foto es de una deportacion de 2018, cuando trump estaba gobernando. & SPA & FALSE \\
Although, to be fair, that photo is from a 2018 deportation, when Trump was in power. & EN &  \\
\hline
la propuesta de k!nadim me parece muy interesante de cara a eurovision, es facilona y divertida \#benidormfestsemi1 & SPA & FALSE \\
The proposal by k!nadim seems very interesting to me for Eurovision; it's easy and fun. \#benidormfestsemi1 & EN &  \\
\hline
\end{tabular}
\caption{Spanish sentences with labels and English translations.}
\label{tab:SampleSpanish}
\end{table}

\end{document}